\documentclass[]{sensenova}

\usepackage{graphicx}
\usepackage{array}
\usepackage{booktabs}
\usepackage{float}
\usepackage{enumitem}
\usepackage{listings}
\usepackage{subcaption}
\usepackage{adjustbox}
\usepackage{mmstyles}

\tcbuselibrary{breakable,skins,listings}

\definecolor{ourscyan}{RGB}{232,248,248}
\definecolor{promptblue}{RGB}{30,92,156}
\definecolor{promptblueback}{RGB}{244,250,255}
\definecolor{findingblue}{HTML}{2B8AC6}

\def\ourmethod{\texttt{CARE}}

\newcommand{\tabcite}[1]{\raisebox{0.15ex}{\mbox{\tiny\citep{#1}}}}
\newcommand{\rlver}{\mbox{RLVER}~\tabcite{wang2026rlver}}
\newcommand{\perm}{\mbox{PERM}~\tabcite{wang2026perm}}
\newcommand{\kardia}{\mbox{KARDIA-R1}~\tabcite{yuan2026kardiar1}}
\newcommand{\oursrlver}{\texttt{CARE(R)}}
\newcommand{\oursmapo}{\texttt{CARE(M)}}
\newcommand{\maporep}{\mbox{MICA (rep.)}~\tabcite{zhang2026mica}}
\newcommand{\qwenbase}{Qwen2.5-7B-Instruct}
\newcommand{\derivedmodel}[1]{\hspace*{1.1em}#1}
\newcommand{\meanstd}[2]{#1\raisebox{-0.32ex}{\scalebox{0.68}{\ensuremath{\pm}\,#2}}}
\newcommand{\best}[1]{\textbf{#1}}
\newcommand{\second}[1]{\underline{#1}}

\newcommand{\principle}[1]{%
    \begin{tcolorbox}[
        enhanced,
        breakable,
        colback=findingblue!6,
        colframe=findingblue!95,
        boxrule=0.75pt,
        arc=4pt,
        boxsep=3pt,
        left=4pt,
        right=4pt,
        top=4pt,
        bottom=4pt,
        before skip=8pt,
        after skip=8pt,
        drop shadow=gray!25!white
    ]
    \noindent #1
    \end{tcolorbox}
}

\newtcblisting{promptbox}[1]{
    enhanced,
    breakable,
    listing only,
    colback=promptblueback,
    colframe=promptblue,
    coltitle=white,
    colbacktitle=promptblue,
    title={#1},
    title after break={#1\ {\normalfont\small(continued)}},
    fonttitle=\bfseries\large,
    boxrule=0.8pt,
    arc=2mm,
    outer arc=2mm,
    left=8pt,
    right=8pt,
    top=7pt,
    bottom=7pt,
    toptitle=4pt,
    bottomtitle=4pt,
    listing options={
        basicstyle=\ttfamily\footnotesize,
        breaklines=true,
        breakatwhitespace=false,
        columns=flexible,
        keepspaces=true,
        showstringspaces=false
    }
}

\title{Evolving Support Priorities in Empathetic \\ Reinforcement Learning}

\author[1,2,3,*]{Pengyu Huang}
\author[1,2,3,*]{Zhiyuan Han}
\author[2,\ddagger]{Wenwen Tong}
\author[2]{Hewei Guo}
\author[2]{Jiangnan Chen}
\author[4]{\\Sirui Chen}
\author[2,\dagger]{Lewei Lu}
\author[1,\dagger]{Beier Zhu}
\author[1,\dagger]{Xun Yang}

\affiliation[1]{University of Science and Technology of China}
\affiliation[2]{SenseTime Research}
\let\firstaffiliationrow\affiliationlist
\renewcommand{\affiliationlist}{}
\affiliation[3]{Institute of Artificial Intelligence, Hefei Comprehensive National Science Center}
\affiliation[4]{Tongji University}
\let\secondaffiliationrow\affiliationlist
\renewcommand{\affiliationlist}{\firstaffiliationrow\\\secondaffiliationrow}

\contribution[*]{Equal contribution}
\contribution[\dagger]{Co-corresponding authors}
\contribution[\ddagger]{Project lead}

\abstract{
We identify a fundamental mismatch in empathetic reinforcement learning:
support priorities evolve with the dialogue state, yet existing methods typically
optimize predefined reward specifications that remain fixed across turns.
To model these \textit{evolving support priorities}, we organize empathetic
support along cognitive, affective, and proactive empathy, and propose
\texttt{C}ontext-\texttt{A}daptive \texttt{R}ubric
\texttt{E}volution~(\ourmethod).
At each turn, \ourmethod~generates a context-adaptive rubric by adjusting both
the weights of these three empathy dimensions and their fine-grained evaluation
criteria.
The rubric generator is trained with turn-level rubric supervision and human
preference data through supervised fine-tuning followed by preference-based
reinforcement learning, and then serves as an adaptive reward interface for
online empathetic RL.
Integrated with both RLVER and MICA, \ourmethod~achieves state-of-the-art
performance across SentientBench, EQBench3, and EMPA under three independent
LLM judges.
Notably, on EMPA, \ourmethod~improves EPM-Idx over the strongest baseline by at least
13 points under all three judges, including an increase from 28.11 to 83.54
under Gemini-2.5-Pro.
Further analyses show that learned rubric priorities vary across
dialogue stages and user emotions, demonstrating \ourmethod~adapts what is
rewarded as support needs evolve.
}

\checkdata[Project page]{\url{https://asa13opp.github.io/CARE-project-page}}

\begin{document}
\maketitle
\enlargethispage{8mm}
\newlength{\regulartextfloatsep}
\setlength{\regulartextfloatsep}{\textfloatsep}
\setlength{\textfloatsep}{0pt}
\AddToHookNext{shipout/after}{\global\textfloatsep=\regulartextfloatsep\relax}

\begin{figure}[!b]
    \centering
    \includegraphics[width=\linewidth]{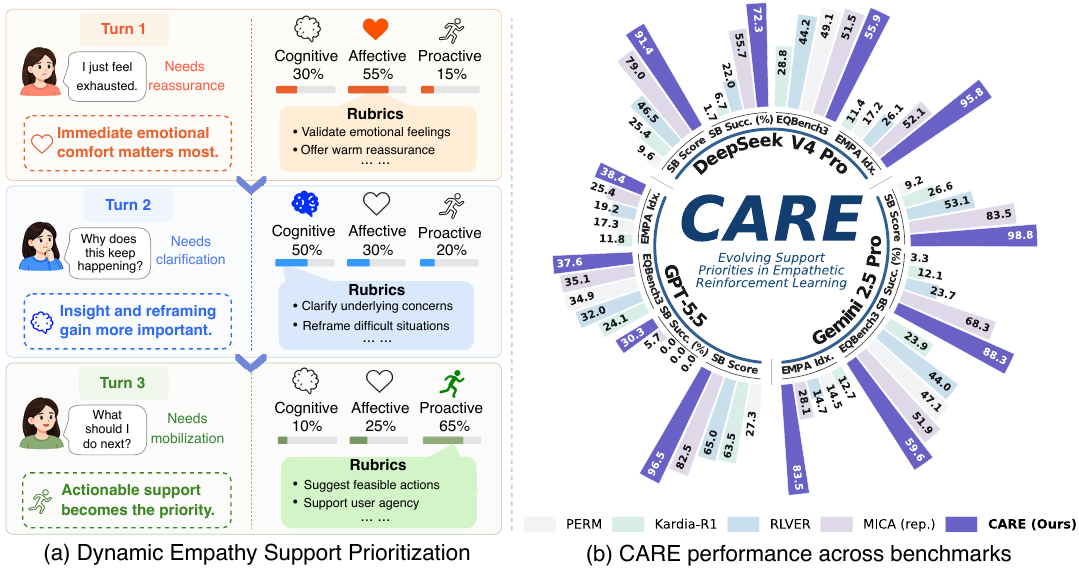}
 \caption{
 {Evolving support priorities in empathetic reinforcement learning.}
\textbf{(a)} 
\ourmethod~constructs a context-adaptive rubric over three empathy dimensions—\textbf{cognitive, affective, and proactive}—with state-dependent weights and fine-grained criteria. 
\textbf{(b)} \ourmethod~consistently outperforms prior methods across benchmarks and judges. Notably, on EMPA~\citep{zhang2026empaevaluatingpersonaalignedempathy}, our \ourmethod~improves EPM-Idx over the strongest baseline with  \textbf{double-digit gains} under all three LLM judges.
}
    \label{fig:teaser}
\vspace{-2mm}
\end{figure}

\section{Introduction}
\label{sec:introduction}

Empathetic dialogue systems aim to understand users’ emotional states and provide appropriate support~\citep{rashkin2019towards,liu_towards_2021}.
Recent work increasingly adopts reinforcement learning to optimize empathetic response quality from feedback~\citep{zhang2026empaevaluatingpersonaalignedempathy,zhang2026mica}.
However, empathetic dialogue is inherently \textbf{non-verifiable}: response quality is open-ended and lacks a unique ground-truth answer, making rule-based rewards difficult to construct.
This has motivated the use of rubric-based evaluators, which decompose response quality into explicit natural-language criteria and translate the resulting judgments into reward signals for empathetic dialogue~\citep{wang2026rlver,yuan2026kardiar1,wang2026perm}.

Existing methods typically use a \textbf{fixed rubric} throughout the dialogue, keeping reward priorities unchanged across turns. However, empathetic support is inherently dynamic: what a user needs can change as a conversation unfolds. As illustrated in Figure~\ref{fig:teaser}(a, left), a single user may initially need reassurance, later seek understanding, and eventually benefit from actionable guidance. This is consistent with psychological research on effective support, which emphasizes matching support to the recipient's current needs and the context of the interaction~\citep{feeney2015new,zaki2020integrating}. These observations motivate the \textit{{Principle of Evolving Support Priorities}}:

\principle{\centering{\textit{Support priorities should adapt to the current dialogue state as it evolves.}}}

To operationalize this principle, we decompose support priorities into three
complementary dimensions following~\citet{zhang2026empaevaluatingpersonaalignedempathy}:
\textbf{cognitive empathy} for understanding and reframing the user's perspective and concerns,
\textbf{affective empathy} for recognizing and validating the user's emotional experience, and
\textbf{proactive empathy} for strengthening the user's agency and ability to move forward (\S\ref{sec:setup}).
Building on this formulation, we propose \texttt{C}ontext-\texttt{A}daptive
\texttt{R}ubric \texttt{E}volution~(\ourmethod), a framework that makes
reward specifications adaptive to the current dialogue state.
As shown in Figure~\ref{fig:teaser}(a, right), \ourmethod~generates a
context-adaptive rubric at each turn by adjusting both the weights of the
three empathy dimensions and their fine-grained evaluation criteria.

To build such a rubric generator, we first construct turn-level rubric supervision from empathetic dialogue data and collect human preference pairs over responses (\S\ref{sec:data}). We then train the generator in two stages: supervised fine-tuning learns the rubric structure and how support priorities vary with context, while preference-based reinforcement learning favors rubrics that better distinguish human-preferred from dispreferred responses (\S\ref{sec:generator}).

Once trained, the rubric generator serves as an adaptive reward interface for online empathetic RL. At each turn, \ourmethod~generates a state-specific rubric from the dialogue context, which a rubric-conditioned evaluator uses to score candidate responses. This interface can be integrated into existing empathetic RL pipelines without changing their rollout or optimization procedures (\S\ref{sec:online}). We apply \ourmethod~to both RLVER~\citep{wang2026rlver} and MICA~\citep{zhang2026mica}, and evaluate the resulting policies with three strong frontier LLM judges from distinct model families---DeepSeek-V4-Pro, Gemini-2.5-Pro, and GPT-5.5---to provide diverse evaluation perspectives and reduce dependence on any single evaluator. Across SentientBench~\citep{zhang2025sentientagentjudgeevaluating}, EQBench3~\citep{eqbench3_repo_2025}, and EMPA~\citep{zhang2026empaevaluatingpersonaalignedempathy}, \ourmethod~establishes a new state of the art under all three judges.  As highlighted in Figure~\ref{fig:teaser}(b), the RLVER-based \oursrlver{} boosts EMPA EPM-Idx from 28.11 to 83.54 under Gemini-2.5-Pro, with double-digit gains over the strongest baseline under the other two judges as well. More broadly, \oursrlver{} and \oursmapo{} rank first and second on every reported aggregate metric under all three judges, showing that context-adaptive rubrics provide a consistently stronger reward interface across distinct empathetic RL pipelines. Further analyses show that rubric priorities shift across dialogue stages and user emotions, demonstrating that \ourmethod~learns to adapt what is rewarded as support needs evolve.

Our contributions are three-fold:
\textbf{(1) Principle \& Data:}
We formulate the \textit{Principle of Evolving Support Priorities}, which calls for reward priorities to adapt with the dialogue state, and operationalize it by constructing turn-level rubric supervision and human preference data.
\textbf{(2) Methodology:}
We propose \ourmethod, which learns a context-adaptive rubric generator through supervised fine-tuning and preference-based reinforcement learning, providing turn-specific reward specifications for online empathetic RL.
\textbf{(3) Performance:}
\ourmethod~establishes a new state of the art across three empathy benchmarks and consistently improves both RLVER and MICA, while analyses show that the learned reward priorities adapt across dialogue stages and user states.

\section{Related Work}
\label{sec:related}

\noindent\textbf{Empathetic dialogue.}
Early empathetic dialogue research mainly improved response generation through emotion modeling and predefined support strategies~\citep{liu_towards_2021,zhang_escot_2024,cheng2022improving}. Such methods typically select strategies within a fixed taxonomy, adapting the response strategy rather than the criteria used to reward it. More recent work has introduced LLM RL and interactive environments for optimizing empathetic behavior over multi-turn conversations. Sentient Agent as a Judge~\citep{zhang2025sentientagentjudgeevaluating} enables user simulation and emotion-based feedback, which RLVER~\citep{wang2026rlver} uses for policy optimization. EMPA~\citep{zhang2026empaevaluatingpersonaalignedempathy} introduces structured empathy states, while MICA~\citep{zhang2026mica} further addresses long-horizon credit assignment. Kardia-R1~\citep{yuan2026kardiar1} and PERM~\citep{wang2026perm} enrich reward modeling with task-specific empathy criteria. Yet support priorities evolve across dialogue turns, while the reward specifications used by these methods remain largely fixed, contrary to the \textit{Principle of Evolving Support Priorities}. \ourmethod~instead adapts the reward specification to the dialogue state, enabling policies to better track changing support priorities.

\noindent\textbf{Rubric-based reward modeling.}
Rubric-based reward modeling provides explicit evaluation criteria for open-ended tasks where exact verification is unavailable. Rubrics as Rewards~\citep{gunjal_rubrics_2025} shows that natural-language rubrics can serve as effective reward signals beyond verifiable domains.
Recent work further learns the rubric itself, making evaluation criteria adaptive rather than fixed. Query-specific rubric learning~\citep{lv2026learning_query_specific_rubrics} learns preference-aligned rubrics for individual queries, PARL~\citep{qiu_preference-aware_2026} induces personalized rubrics from user histories, and DynamicRubric~\citep{wang2026co} co-evolves rubrics with the policy by conditioning evaluation criteria on current candidate response sets. Collectively, these works demonstrate that rubric learning can adapt evaluation standards beyond fixed templates. \ourmethod~targets a different source of variation, adapting the reward specification to the evolving dialogue state \textit{rather than to the task or current candidate set}. Within a single empathetic dialogue, the appropriate support objective can change from one turn to the next.

\section{Method}
\label{sec:method}

\begin{figure}[t]
    \centering
    \includegraphics[width=\linewidth]{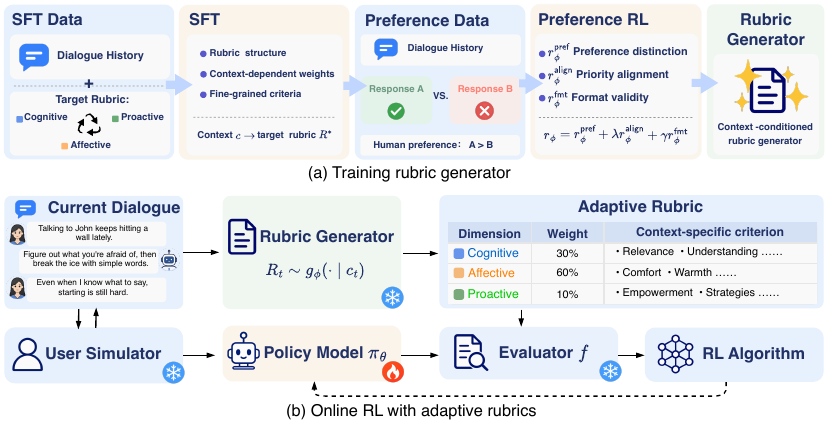}
    \caption{Overview of \ourmethod. Turn-level supervision and human response preferences train a context-conditioned rubric generator.  The frozen generator then
    supplies adaptive evaluation criteria for online dialogue RL,
    with one rubric shared by candidate responses from the same context.}
    \label{fig:method-overview}
\end{figure}

As summarized in Figure~\ref{fig:method-overview}, \ourmethod~learns
context-adaptive reward specifications for individual dialogue turns and uses
them to optimize the dialogue policy.
\S\ref{sec:setup} introduces the dialogue setting and adaptive rubric
formulation.
\S\ref{sec:data} constructs turn-level rubric supervision and human
preference data.
\S\ref{sec:generator} trains the rubric generator through SFT and
preference-based reinforcement learning, and \S\ref{sec:online} uses the
trained generator to provide evolving reward specifications during online
dialogue RL.

\subsection{Setup}
\label{sec:setup}

We organize empathetic support into three dimensions,
$\mathcal D=\{\mathsf{Cog},\mathsf{Aff},\mathsf{Pro}\}$, corresponding to
cognitive, affective, and proactive empathy, following~\citet{zhang2026empaevaluatingpersonaalignedempathy,zhang2026mica}.
This decomposition draws on psychological distinctions among cognitive
understanding, affective experience, and the motivation to improve others'
well-being~\citep{zaki2020integrating}.
These three dimensions are defined as follows:

\begin{itemize}[leftmargin=1.5em, itemsep=0.3em, topsep=0.4em]
    \item \textbf{Cognitive Empathy:}
    accurately understanding and structuring the user's mental state, going beyond restatement to clarify or reframe how the user understands the situation.

    \item \textbf{Affective Empathy:}
    recognizing and validating the user's emotional experience, while helping them feel seen and accepted rather than relying on exaggerated or formulaic displays of empathy.

    \item \textbf{Proactive Empathy:}
    building on understanding and validation to strengthen the user's agency and ability to move forward, rather than merely offering advice or generic encouragement.
\end{itemize}

Effective empathetic support requires \textit{not only covering all three dimensions, but also adapting their priorities to the current dialogue state.}
For example, early in a distressing conversation, affective empathy is prioritized; as the user begins to articulate the underlying conflict, cognitive empathy becomes important; and when the user turns to what to do next, proactive empathy takes priority.

Formally, at assistant turn $t$, the policy conditions on the dialogue context
$c_t=(u_1,y_1,\ldots,u_t)$, where $u_t$ is the current user utterance, and
generates the assistant response
$y_t\sim\pi_\theta(\cdot\mid c_t)$.
At each turn, \ourmethod~uses a generator $g_\phi$ to produce a
context-dependent rubric:
\begin{equation}
    \mathcal R_t=
    \{(w_{t,d},\mathcal C_{t,d})\}_{d\in\mathcal D} \sim g_\phi(\cdot\mid c_t),
    \qquad
    \mathcal C_{t,d}
    =
    \{\kappa_{t,d,i}\}_{i=1}^{|\mathcal C_{t,d}|},
    \qquad
    w_{t,d}\geq 0,\ 
    \sum_{d\in\mathcal D} w_{t,d}=1,
    \label{eq:rubric}
\end{equation}
where $w_{t,d}$ denotes the priority weight for dimension $d\in\mathcal D$, while
$\kappa_{t,d,i}\in\mathcal C_{t,d}$ denotes its $i$-th fine-grained evaluation
criterion.
The three empathy dimensions remain fixed across turns, while their weights
and fine-grained criteria adapt to the current dialogue context.

\subsection{Turn-level rubric data construction}
\label{sec:data}

\noindent\textbf{SFT data construction.}
We construct turn-level rubric targets from emotional-support dialogues in
MINT~\citep{zhan_discourse_2026} and ESConv~\citep{liu_towards_2021}.
For each assistant turn $t$, the datasets provide the dialogue context $c_t$,
reference response $y_t^\star$, and the support-strategy annotations.
These strategies 
are mapped to the three empathy dimensions $\mathcal D$ through a fixed
mapping $m$ (details in \S\ref{app:strategy-dimension-mapping}), providing coarse supervision for turn-specific support priorities. Conditioned on $c_t$, $y_t^\star$, support-strategies, and the mapped dimension annotations, an LLM generates the target rubric
$\mathcal R_t^\star$.  We further filter generated rubrics using two independent judges for rubric quality. This yields 1.5K SFT examples of the form $(c_t,\mathcal R_t^\star)$.
The complete construction and filtering procedure are in
\S\ref{app:sft-data}.

\noindent\textbf{Preference RL data construction.}
We construct preference data from collected multi-turn emotional-support
conversations with human users. From each dialogue, we select a preceding
context $c$ and roll out multiple strong LLMs from the same context to obtain
comparable candidate responses. The candidates are anonymized and presented
in randomized pairs to human annotators, with three judgments per pair and
majority vote determining the preferred response $y^+$ and the dispreferred
response $y^-$. This yields 2.8K preference pairs $\{(c,y^+,y^-)\}$.

For each preferred response $y^+$, we further obtain support-strategy annotations
$\mathcal A^+$ using the MINT annotator and map them to the three empathy
dimensions using $m:\mathcal A\rightarrow\mathcal D$.
We denote the resulting strategy count for dimension $d$ as
\begin{equation}
n_d^+
=
\left|
\{a\in\mathcal A^+ : m(a)=d\}
\right|.    
\end{equation}
These counts provide a coarse signal of the support priorities expressed in
$y^+$ and are used for priority alignment during generator training. Full details are provided in
\S\ref{app:preference-data}.

\subsection{Learning the rubric generator}
\label{sec:generator}

We train the rubric generator $g_\phi$ in two stages. We first perform standard SFT to teach the generator to construct a rubric from the dialogue context, including its dimension priorities and evaluation criteria. We then apply preference-based RL to align the generated rubrics with human feedback.

Given a preference example $(c,y^+,y^-)$\footnote{We omit turn index $t$ for the ease of readability.}, we sample a rubric
$\mathcal R=\{(w_{d},\mathcal C_{d})\}_{d\in\mathcal D}\sim g_\phi(\cdot\mid c)$. For each criterion $\kappa_{d,i}\in\mathcal C_d$, an LLM judge $f$ assigns
scores to the preferred and dispreferred responses, respectively. We define the preference reward as
\begin{equation}
s^\pm_{d,i}
=
f(c,y^\pm,\kappa_{d,i}),
\qquad
r_{\phi}^{\mathsf{pref}}
=
\sum_{d\in\mathcal D}
w_d
\sum_{i=1}^{|\mathcal C_d|}
\tfrac{1}{|\mathcal C_d|}
\tanh\left(
{(s^+_{d,i}-s^-_{d,i})}/{\tau}
\right),
\label{eq:pref}
\end{equation}
where $\tau \in \mathbb{R}_+$ controls score sensitivity.
The reward is higher when the rubric favors the preferred response over the dispreferred one.

However, $r^{\mathsf{pref}}_\phi$ only encourages the rubric to score the preferred response higher than the dispreferred one; \textit{it does not ensure that the dimension weights reflect the support strategies in $y^+$.}
We therefore introduce a priority-alignment reward $r^{\mathsf{align}}_\phi$ based on the strategy counts $\{n_d^+\}_{d\in\mathcal D}$.
Specifically, the dimension with the largest $n_d^+$ should receive the highest
rubric weight, while a dimension with $n_d^+=0$ should not outweigh any
dimension with $n_d^+>0$.
$r^{\mathsf{align}}_\phi$ is $1$ when both constraints are satisfied, decreases by
$0.5$ for each violation, and is lower-bounded at $0$.
The alignment reward supervises the ranking of dimension weights rather than their exact values.

We additionally use a binary format reward $r^{\mathsf{fmt}}_\phi$, which is $1$
when the generated rubric satisfies the required format and $0$ otherwise.
The complete generator reward is
\begin{equation}
r_\phi
=
r^{\mathsf{pref}}_\phi
+\lambda r^{\mathsf{align}}_\phi
+\gamma r^{\mathsf{fmt}}_\phi,
\label{eq:generator_reward}
\end{equation}
where $\lambda \in \mathbb{R}_+$ and $\gamma \in \mathbb{R}_+$ are weighting hyperparameters. The generator 
$g_\phi$ is optimized using standard GRPO~\citep{shao2024grpodeepseek} with $r_\phi$.

\subsection{Online RL with evolving rubrics}
\label{sec:online}

We use the trained rubric generator $g_\phi$ and LLM judge $f$ as the reward
interface for online dialogue RL. At assistant turn $t$, the generator produces
a rubric $\mathcal R_t\sim g_\phi(\cdot\mid c_t)$ from the current 
context. The policy $\pi_\theta$ then samples a group of $M$ responses
$\{y_{t,j}\}_{j=1}^{M}\sim\pi_\theta(\cdot\mid c_t)$ for GRPO optimization.
For clarity, we omit the turn index $t$ and response index $j$ below.
For each criterion $\kappa_{d,i}$, the evaluator $f$ produces a normalized
score $s_{d,i}\in[0,1]$. The turn-specific reward for training policy $\pi_\theta$ is
\begin{equation}
r_\theta
=
\sum_{d\in\mathcal D}
w_d
\sum_{i=1}^{|\mathcal C_d|}
\tfrac{1}{|\mathcal C_d|}
s_{d,i}.
\label{eq:policy_reward}
\end{equation}
The adaptive reward \(r_\theta\) directly replaces the original reward in RLVER and MICA. Since all responses within a rollout group share the same dialogue context, they are evaluated under the same rubric, while a new rubric is generated as the context changes across turns.

\section{Experiments}
\label{sec:experiments}

\begin{table}[!t]
    \centering
    \setlength{\belowcaptionskip}{6pt}
    \caption{\textbf{Main comparison across SentientBench, EQBench3, and EMPA under three LLM judges.} Additional results for EQBench3 and EMPA are provided in \S\ref{app:detailed-results}. Standard deviations  are computed over three random seeds; bold and underline mark the best and second-best results.}
    \label{tab:main-results}
    \setlength{\tabcolsep}{0pt}
    \small
    \begin{tabular}{@{}p{0.28\linewidth}>{\centering\arraybackslash}p{0.144\linewidth}>{\centering\arraybackslash}p{0.144\linewidth}>{\centering\arraybackslash}p{0.144\linewidth}>{\centering\arraybackslash}p{0.144\linewidth}>{\centering\arraybackslash}p{0.144\linewidth}@{}}
        \toprule
        \multicolumn{1}{c}{} & \multicolumn{3}{c}{\textbf{SentientBench}} & \textbf{EQBench3} & \textbf{EMPA} \\
        \cmidrule(lr){2-4} \cmidrule(lr){5-5} \cmidrule(lr){6-6}
        \textbf{Model} & \textbf{Score} $\boldsymbol{\uparrow}$ & \textbf{Succ. (\%)} $\boldsymbol{\uparrow}$ & \textbf{Fail (\%)} $\boldsymbol{\downarrow}$ & \textbf{Overall} $\boldsymbol{\uparrow}$ & \textbf{EPM-Idx} $\boldsymbol{\uparrow}$ \\
        \midrule
        \multicolumn{6}{@{}l}{\textit{Judge: DeepSeek-V4-Pro}} \\
        \qwenbase & \meanstd{9.14}{1.48} & \meanstd{3.33}{0.94} & \meanstd{87.67}{0.47} & \meanstd{42.53}{2.18} & \meanstd{15.38}{1.06} \\
        \derivedmodel{\perm} & \meanstd{9.57}{0.62} & \meanstd{1.67}{0.94} & \meanstd{86.67}{1.25} & \meanstd{49.07}{1.03} & \meanstd{17.21}{1.17} \\
        \derivedmodel{\kardia} & \meanstd{25.45}{2.46} & \meanstd{6.67}{1.70} & \meanstd{58.33}{2.87} & \meanstd{28.75}{1.21} & \meanstd{11.42}{0.34} \\
        \derivedmodel{\rlver} & \meanstd{46.53}{2.13} & \meanstd{22.00}{0.00} & \meanstd{30.33}{4.50} & \meanstd{44.23}{2.23} & \meanstd{26.05}{3.37} \\
        \derivedmodel{\maporep} & \meanstd{78.98}{1.39} & \meanstd{55.67}{1.25} & \meanstd{7.00}{0.82} & \meanstd{51.47}{0.58} & \meanstd{52.10}{4.12} \\
        \rowcolor{ourscyan}\derivedmodel{\oursmapo} & \second{\meanstd{87.40}{0.86}} & \second{\meanstd{62.67}{1.70}} & \second{\meanstd{3.33}{0.47}} & \best{\meanstd{57.53}{1.20}} & \second{\meanstd{94.84}{1.08}} \\
        \rowcolor{ourscyan}\derivedmodel{\oursrlver} & \best{\meanstd{91.41}{2.09}} & \best{\meanstd{72.33}{2.87}} & \best{\meanstd{2.00}{0.82}} & \second{\meanstd{55.88}{1.59}} & \best{\meanstd{95.77}{0.33}} \\
        \midrule
        \multicolumn{6}{@{}l}{\textit{Judge: Gemini-2.5-Pro}} \\
        \qwenbase & \meanstd{7.55}{1.85} & \meanstd{1.67}{0.47} & \meanstd{91.00}{2.16} & \meanstd{43.43}{1.74} & \meanstd{15.95}{1.87} \\
        \derivedmodel{\perm} & \meanstd{9.25}{1.12} & \meanstd{3.33}{1.70} & \meanstd{88.00}{2.16} & \meanstd{47.13}{2.63} & \meanstd{14.53}{1.03} \\
        \derivedmodel{\kardia} & \meanstd{26.65}{1.61} & \meanstd{12.06}{1.19} & \meanstd{66.55}{1.32} & \meanstd{23.95}{0.36} & \meanstd{12.72}{0.91} \\
        \derivedmodel{\rlver} & \meanstd{53.12}{1.18} & \meanstd{23.67}{2.05} & \meanstd{29.33}{0.94} & \meanstd{43.98}{1.23} & \meanstd{14.65}{0.43} \\
        \derivedmodel{\maporep} & \meanstd{83.54}{0.32} & \meanstd{68.33}{3.30} & \meanstd{10.67}{1.25} & \meanstd{51.93}{1.05} & \meanstd{28.11}{2.28} \\
        \rowcolor{ourscyan}\derivedmodel{\oursmapo} & \second{\meanstd{97.76}{0.79}} & \second{\meanstd{85.33}{2.87}} & \second{\meanstd{0.67}{0.47}} & \second{\meanstd{58.82}{2.94}} & \second{\meanstd{81.20}{4.81}} \\
        \rowcolor{ourscyan}\derivedmodel{\oursrlver} & \best{\meanstd{98.85}{0.42}} & \best{\meanstd{88.33}{2.87}} & \best{\meanstd{0.00}{0.00}} & \best{\meanstd{59.60}{3.94}} & \best{\meanstd{83.54}{3.94}} \\
        \midrule
        \multicolumn{6}{@{}l}{\textit{Judge: GPT-5.5}} \\
        \qwenbase & \meanstd{31.94}{0.79} & \meanstd{0.00}{0.00} & \meanstd{36.33}{1.25} & \meanstd{32.10}{0.18} & \meanstd{22.97}{2.75} \\
        \derivedmodel{\perm} & \meanstd{27.31}{0.59} & \meanstd{0.00}{0.00} & \meanstd{45.00}{4.24} & \meanstd{34.88}{0.58} & \meanstd{17.29}{2.90} \\
        \derivedmodel{\kardia} & \meanstd{63.54}{0.70} & \meanstd{0.00}{0.00} & \meanstd{2.67}{0.94} & \meanstd{24.15}{0.74} & \meanstd{11.76}{0.19} \\
        \derivedmodel{\rlver} & \meanstd{64.96}{2.46} & \meanstd{0.00}{0.00} & \meanstd{1.33}{0.94} & \meanstd{32.00}{0.61} & \meanstd{19.19}{1.88} \\
        \derivedmodel{\maporep} & \meanstd{82.51}{0.92} & \meanstd{5.67}{1.25} & \meanstd{0.67}{0.47} & \meanstd{35.10}{0.39} & \meanstd{25.42}{1.41} \\
        \rowcolor{ourscyan}\derivedmodel{\oursmapo} & \second{\meanstd{94.84}{0.31}} & \second{\meanstd{27.67}{5.44}} & \best{\meanstd{0.00}{0.00}} & \second{\meanstd{37.58}{0.93}} & \second{\meanstd{36.70}{4.38}} \\
        \rowcolor{ourscyan}\derivedmodel{\oursrlver} & \best{\meanstd{96.50}{0.47}} & \best{\meanstd{30.33}{2.62}} & \best{\meanstd{0.00}{0.00}} & \best{\meanstd{37.65}{0.89}} & \best{\meanstd{38.44}{3.96}} \\
        \bottomrule
    \end{tabular}
    \vspace{-6pt}
\end{table}

We first summarize three key empirical findings that characterize the strengths of \ourmethod:

\noindent\textbf{Substantial gains over prior empathetic dialogue methods.}
     E.g., with DeepSeek-V4-Pro as the judge, \oursrlver{} achieves
91.41 on SentientBench, 55.88 on EQBench3, and 95.77 on EMPA,
improving over the strongest prior methods by 12.43, 4.41, and 43.67 points,
respectively (Tab.~\ref{tab:main-results}).

   \noindent \textbf{Competitive performance against frontier LLMs.}
    E.g., on EQBench3,
    \ourmethod~with Qwen3-32B reaches 76.35, surpassing GLM-5.2-753B
    (74.90) and approaching DeepSeek-V4-Pro-1.6T (77.75), despite using a
    substantially smaller backbone (Tab.~\ref{tab:eqbench3-scale-ablation}).

   \noindent\textbf{Interpretable adaptation with broadly preserved general capabilities.}
    \ourmethod~does not show systematic degradation on coding, mathematical
    reasoning, or instruction following. Meanwhile, visualizations of learned
    dimension weights and turn-specific rubrics reveal clear state-dependent
    patterns across dialogue turns and user emotions (\S\ref{sec:analysis}).

\subsection{Setup}
\label{sec:exp-setup}

\noindent\textbf{Baselines.}
We consider two comparison settings.
For the controlled comparison with prior empathetic dialogue methods in Table~\ref{tab:main-results}, we compare against PERM~\citep{wang2026perm}, Kardia-R1~\citep{yuan2026kardiar1}, RLVER~\citep{wang2026rlver}, and our reproduction of MICA~\citep{zhang2026mica}, with all methods using Qwen2.5-7B-Instruct~\citep{yang2024qwen25} as the common policy backbone $\pi_\theta$.
For the comparison with frontier LLMs in Table~\ref{tab:eqbench3-scale-ablation}, we apply \ourmethod~to Qwen3 models~\citep{yang2025qwen3} ranging from 8B to 32B and compare their performance with strong frontier LLMs.
We integrate \ourmethod~into RLVER and MICA by replacing their original rewards with our adaptive rubric-based reward, yielding \oursrlver{} and \oursmapo{}, respectively.
For reward computation, DeepSeek-V4-Flash serves as the rubric-conditioned evaluator $f$, while the rubric generator $g_\phi$ is built on Qwen3-8B and remains frozen throughout online policy optimization.

\noindent\textbf{Evaluation.}
We evaluate on three empathy benchmarks. SentientBench~\citep{zhang2025sentientagentjudgeevaluating} assesses social cognition through multi-turn supportive dialogues. EQBench3~\citep{eqbench3_repo_2025} probes emotional intelligence through role-play and transcript analysis. EMPA~\citep{zhang2026empaevaluatingpersonaalignedempathy} evaluates persona-aligned empathy. To reduce reliance on a single evaluator, we use DeepSeek-V4-Pro, Gemini-2.5-Pro, and GPT-5.5 as independent judges for all benchmarks. 


\subsection{Main Results}
\label{sec:main-results}
Table~\ref{tab:main-results} shows that \ourmethod~achieves state-of-the-art performance in every evaluated setting and substantially improves both RL pipelines. 

\begin{table}[!t]
    \centering
    \setlength{\belowcaptionskip}{6pt}
    \caption{\textbf{Scaling and frontier comparison on EQBench3 under DeepSeek-V4-Pro.} We compare Base and \ourmethod-enhanced Qwen3 models from 8B to 32B, together with strong frontier LLMs.}
    \label{tab:eqbench3-scale-ablation}
    \small
    \renewcommand{\arraystretch}{0.78}
    \setlength{\tabcolsep}{0pt}
    \begin{tabular}{@{}>{\raggedright\arraybackslash}p{0.23\linewidth}*{8}{>{\centering\arraybackslash}p{0.09625\linewidth}}@{}}
        \toprule
        \textbf{Model} & \textbf{Overall} & \textbf{DoI} & \textbf{ER} & \textbf{DE} & \textbf{WRM} & \textbf{HL} & \textbf{PEI} & \textbf{SD} \\
        \midrule
        GPT-5.5 & 83.35 & 17.57 & 16.98 & 16.64 & 14.32 & 16.72 & 16.72 & 16.16 \\
        Gemini-2.5-Pro & 82.65 & 17.71 & 16.73 & 17.00 & 15.31 & 17.31 & 15.81 & 15.85 \\
        Claude-Sonnet-4.6 & 79.30 & 17.52 & 16.48 & 15.38 & 13.35 & 15.54 & 14.00 & 13.88 \\
        DeepSeek-V4-Pro-1.6T & 77.75 & 16.91 & 16.14 & 15.69 & 13.73 & 16.31 & 13.96 & 13.69 \\
        Qwen3.7-Max & 76.60 & 17.00 & 15.51 & 15.65 & 13.23 & 14.92 & 14.58 & 13.77 \\
        GLM-5.2-753B & 74.90 & 16.84 & 15.43 & 14.62 & 12.46 & 15.12 & 13.35 & 12.92 \\
        \midrule
        \multicolumn{9}{@{}l}{\textit{Qwen3-8B}} \\
        \derivedmodel{Base} & \meanstd{61.60}{0.97} & \meanstd{13.72}{0.25} & \meanstd{12.87}{0.18} & \meanstd{12.56}{0.12} & \meanstd{11.04}{0.04} & \meanstd{11.67}{0.72} & \meanstd{11.01}{0.63} & \meanstd{9.67}{0.64} \\
        \rowcolor{ourscyan}\derivedmodel{\oursrlver} & \meanstd{68.22}{1.59} & \meanstd{14.47}{0.29} & \meanstd{14.09}{0.20} & \meanstd{14.25}{0.57} & \meanstd{12.37}{0.35} & \meanstd{13.55}{0.10} & \meanstd{12.99}{0.72} & \meanstd{12.11}{0.72} \\
        \midrule
        \multicolumn{9}{@{}l}{\textit{Qwen3-14B}} \\
        \derivedmodel{Base} & \meanstd{68.88}{0.93} & \meanstd{15.28}{0.17} & \meanstd{14.26}{0.26} & \meanstd{14.06}{0.37} & \meanstd{12.72}{0.02} & \meanstd{13.82}{0.42} & \meanstd{12.57}{0.46} & \meanstd{11.79}{0.41} \\
        \rowcolor{ourscyan}\derivedmodel{\oursrlver} & \meanstd{73.98}{1.73} & \meanstd{15.88}{0.28} & \meanstd{15.24}{0.36} & \meanstd{15.11}{0.34} & \meanstd{13.46}{0.11} & \meanstd{14.97}{0.23} & \meanstd{14.08}{0.47} & \meanstd{13.37}{0.46} \\
        \midrule
        \multicolumn{9}{@{}l}{\textit{Qwen3-32B}} \\
        \derivedmodel{Base} & \meanstd{73.50}{1.34} & \meanstd{16.07}{0.22} & \meanstd{14.89}{0.34} & \meanstd{14.50}{0.24} & \meanstd{12.72}{0.66} & \meanstd{14.85}{0.55} & \meanstd{14.04}{0.04} & \meanstd{13.23}{0.35} \\
        \rowcolor{ourscyan}\derivedmodel{\oursrlver} & \meanstd{76.35}{2.50} & \meanstd{16.35}{0.46} & \meanstd{15.99}{0.83} & \meanstd{14.76}{0.10} & \meanstd{13.03}{0.45} & \meanstd{15.38}{0.70} & \meanstd{14.41}{0.48} & \meanstd{13.80}{0.72} \\
        \bottomrule
    \end{tabular}
\end{table}

\begin{table}[t]
\centering

\begin{minipage}[t]{0.57\linewidth}
    \captionsetup{
        justification=raggedright,
        singlelinecheck=false,
        skip=5pt
    }
    \captionof{table}{Rubric-design ablation for policy optimization on SentientBench under DeepSeek-V4-Pro.}
    \label{tab:policy-rubric-ablation}

    \small
    \setlength{\tabcolsep}{0pt}
    \begin{tabular}{@{}
        >{\raggedright\arraybackslash}p{0.46\linewidth}
        >{\centering\arraybackslash}p{0.18\linewidth}
        >{\centering\arraybackslash}p{0.18\linewidth}
        >{\centering\arraybackslash}p{0.18\linewidth}
        @{}}
        \toprule
        \textbf{Reward / rubric variant}
        & \textbf{Score} $\boldsymbol{\uparrow}$
        & \textbf{Succ.} $\boldsymbol{\uparrow}$
        & \textbf{Fail} $\boldsymbol{\downarrow}$ \\
        \midrule
        Scalar reward model
        & \meanstd{22.92}{2.56}
        & \meanstd{1.33}{0.47}
        & \meanstd{65.00}{4.32} \\
        SFT-only rubric generator
        & \meanstd{28.21}{4.06}
        & \meanstd{10.67}{3.30}
        & \meanstd{62.00}{2.94} \\
        Uniform rubric criteria
        & \meanstd{41.62}{4.85}
        & \meanstd{1.33}{1.89}
        & \meanstd{41.33}{7.59} \\
        Uniform dimension weights
        & \meanstd{64.20}{0.15}
        & \meanstd{34.33}{1.89}
        & \meanstd{24.00}{1.41} \\
        \rowcolor{ourscyan}
        Full \ourmethod~rubric
        & \best{\meanstd{91.41}{2.09}}
        & \best{\meanstd{72.33}{2.87}}
        & \best{\meanstd{2.00}{0.82}} \\
        \bottomrule
    \end{tabular}
\end{minipage}
\hfill
\begin{minipage}[t]{0.41\linewidth}
    \captionsetup{
        justification=raggedright,
        singlelinecheck=false,
        skip=5pt
    }
\captionof{table}{Ablation of rubric generator training on preference recovery.}
    \label{tab:rubric-generator-ablation}

    \small
    \setlength{\tabcolsep}{0pt}
    \begin{tabular}{@{}
        >{\centering\arraybackslash}p{0.10\linewidth}
        >{\centering\arraybackslash}p{0.10\linewidth}
        >{\centering\arraybackslash}p{0.25\linewidth}
        >{\centering\arraybackslash}p{0.18\linewidth}
        >{\centering\arraybackslash}p{0.18\linewidth}
        >{\centering\arraybackslash}p{0.19\linewidth}
        @{}}
        \toprule
        \textbf{SFT} & \textbf{RL} & \textbf{Reward}
        & \textbf{Win} & \textbf{Tie} & \textbf{Loss} \\
        \midrule
        -- & -- & -- & 49.3 & 13.8 & 36.8 \\
        $\surd$ & -- & -- & 47.6 & 24.1 & 28.3 \\
        -- & $\surd$ & Full & 53.3 & 18.7 & 28.0 \\
        $\surd$ & $\surd$ & w/o $R_{\mathsf{align}}$
        & 56.4 & 9.7 & 33.9 \\
        \rowcolor{ourscyan}
        $\surd$ & $\surd$ & Full
        & \textbf{59.4} & 11.5 & 29.1 \\
        \bottomrule
    \end{tabular}
\end{minipage}

\vspace{-4mm}
\end{table}

\noindent\textbf{SentientBench.}
\ourmethod~delivers substantial gains on this interactive benchmark, improving both dialogue scores and successful support outcomes. The gap is especially pronounced under GPT-5.5, the strictest judge in terms of success rate, where \textbf{\oursrlver{} achieves 30.33\% success compared with only 5.67\% for MICA}, the strongest baseline, while reducing the failure rate to 0\%. 
Notably, RLVER originally trails MICA substantially on SentientBench, but this gap largely disappears after applying \ourmethod: \oursrlver{} matches or slightly outperforms \oursmapo{} on most metrics. This suggests that adaptive reward specifications can matter as much as the choice of the RL pipeline.

\noindent\textbf{EQBench3.}
The gains extend to EQBench3, which evaluates emotional intelligence through role-play and transcript analysis. Across the three judges, the best \textbf{\ourmethod~variant improves Overall by 2.55--7.67 points over the strongest baseline}, while \oursrlver{} and \oursmapo{} consistently rank first and second. These results show that the benefit of adaptive reward specifications extends beyond interactive user-simulator benchmarks.  Complete results are provided in \S~\ref{app:detailed-results}, Table~\ref{tab:eqbench3-full}.

\noindent\textbf{EMPA.}
\ourmethod~shows particularly large gains on EMPA, which evaluates persona-aligned empathy over complete dialogue trajectories. Under DeepSeek-V4-Pro, \oursrlver{} improves EPM-Idx from 52.10 for the strongest baseline MICA to 95.77. Under Gemini-2.5-Pro, the gain is even larger, with EPM-Idx increasing from 28.11 to 83.54---\textbf{nearly three times the baseline}. Across all three judges, \oursrlver{} and \oursmapo{} consistently occupy the top two positions, indicating that the improvement is robust across both evaluator choices and user-simulator-based RL pipelines. Importantly, EPM-Idx jointly reflects outcome quality, process efficiency, and strategic stability (details in \S\ref{app:detailed-results}, Table~\ref{tab:empa-outcome-full}), and the component-level results show particularly strong gains in outcome quality and stability. 

\noindent\textbf{Human evaluation.}
Fifteen annotators with psychology-related backgrounds select the best response among \oursrlver{}, MICA, and RLVER for randomly sampled dialogue contexts, with model identities hidden.
\oursrlver{} is also preferred by human evaluators, receiving 51.5\% of selections, compared with 23.9\% for MICA and 24.6\% for RLVER. See Appendix~\ref{app:human_eval} for details.

\noindent\textbf{Scaling and comparison with frontier LLMs.} Table~\ref{tab:eqbench3-scale-ablation} further evaluates \ourmethod~across Qwen3 model scales and compares the resulting models with strong frontier LLMs. \ourmethod~consistently improves EQBench3 Overall and all other 7 metrics. Notably, \oursrlver{} at 14B already surpasses the 32B Base model (73.98 vs.\ 73.50). At 32B, \oursrlver{} further \textbf{exceeds GLM-5.2}, a 744B-parameter MoE model with 40B active parameters (76.35 vs.\ 74.90), and nearly \textbf{matches Qwen3.7-Max} (76.60). These results demonstrate that the gains from \ourmethod~persist across model scales and substantially narrow the gap between compact Qwen3 models and stronger frontier systems.

\subsection{Ablation studies}
\label{sec:ablation}

\noindent\textbf{Effect of rubric design during policy optimization.}
We isolate the contribution of different rubric components on SentientBench. The scalar reward model variant replaces adaptive rubric scoring with the reward model of MICA \citep{zhang2026mica}, while the SFT-only variant uses the rubric generator before preference-based RL. To separately test the two adaptive components of \ourmethod, Uniform dimension weights retain the generated criteria but set all dimension weights to $1/3$, whereas Uniform rubric criteria retain the generated weights but replace the turn-specific criteria with a fixed six-criterion set containing two representative criteria per empathy dimension. Table~\ref{tab:policy-rubric-ablation} shows that removing either adaptive weights or turn-specific criteria causes substantial performance degradation, while the full \ourmethod~rubric achieves the best results across all metrics. These results confirm that both context-dependent priorities and fine-grained criteria are essential to the effectiveness of the adaptive reward specification.

\noindent\textbf{Ablation of rubric-generator training.}
We evaluate different training variants on 166 held-out chosen--rejected pairs, where the generator receives only the dialogue context and DeepSeek-V4-Flash scores both responses under the generated rubric. As shown in Table~\ref{tab:rubric-generator-ablation}, SFT alone achieves a 47.6\% chosen-response win rate, while adding preference-based RL increases it to 59.4\%. Removing \(r^{\mathsf{align}}_\phi\) reduces the win rate to 56.4\%, showing that both preference discrimination and priority alignment contribute to learning rubrics that better recover human preferences.


\section{Analyses}
\label{sec:analysis}

\begin{figure}[!t]
\centering

\begin{minipage}[t]{0.48\linewidth}
    \vspace{0pt}
    \centering
    \includegraphics[width=\linewidth]{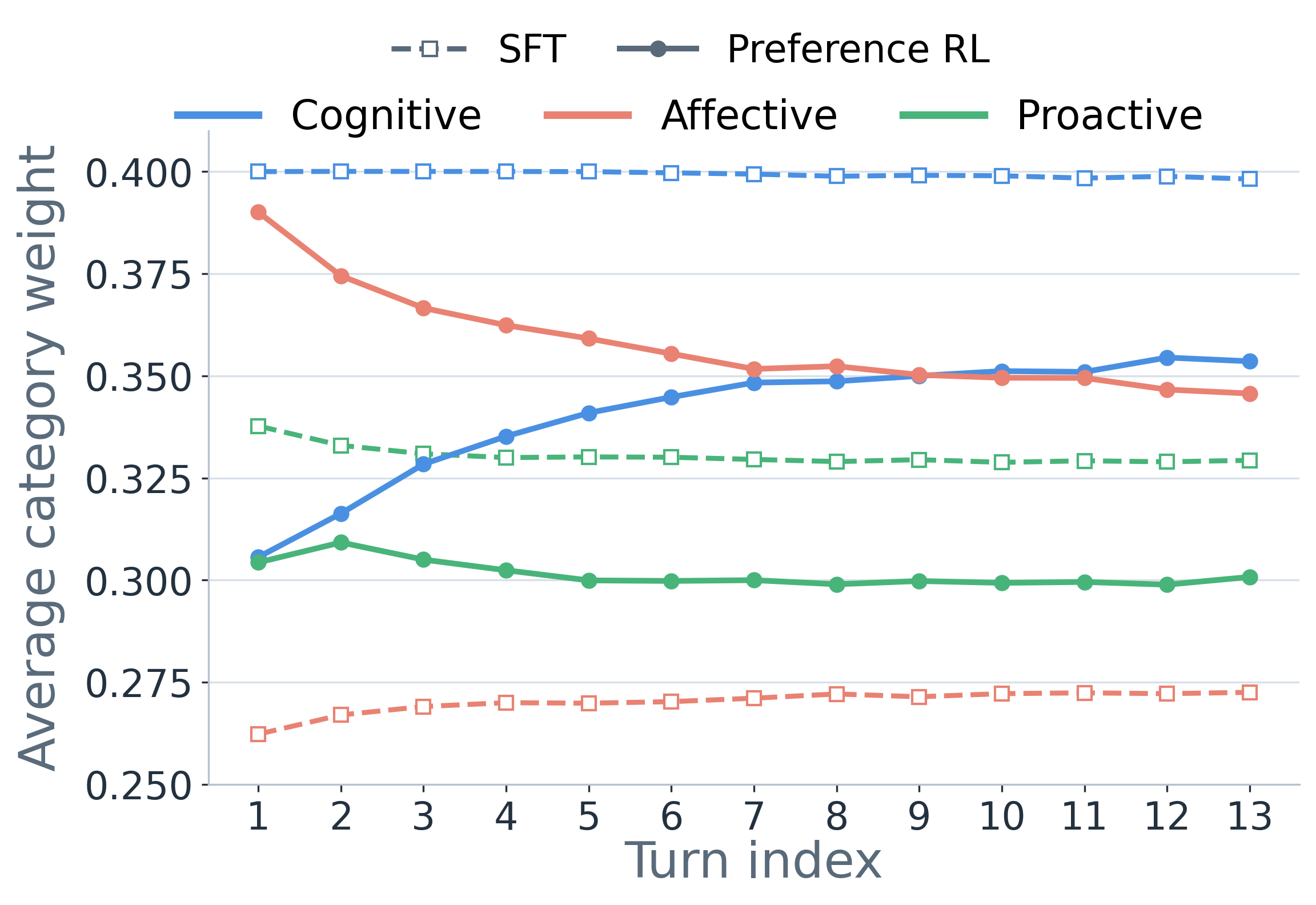}
    \captionof{figure}{
        \textbf{Mean dimension weights }across dialogue turns for the SFT generator
        (dotted) and the preference-RL generator (solid).
    }
    \label{fig:rubric-turn-dynamics}
\end{minipage}
\hfill
\begin{minipage}[t]{0.50\linewidth}
    \vspace{0pt}
    \centering
    \includegraphics[width=\linewidth]{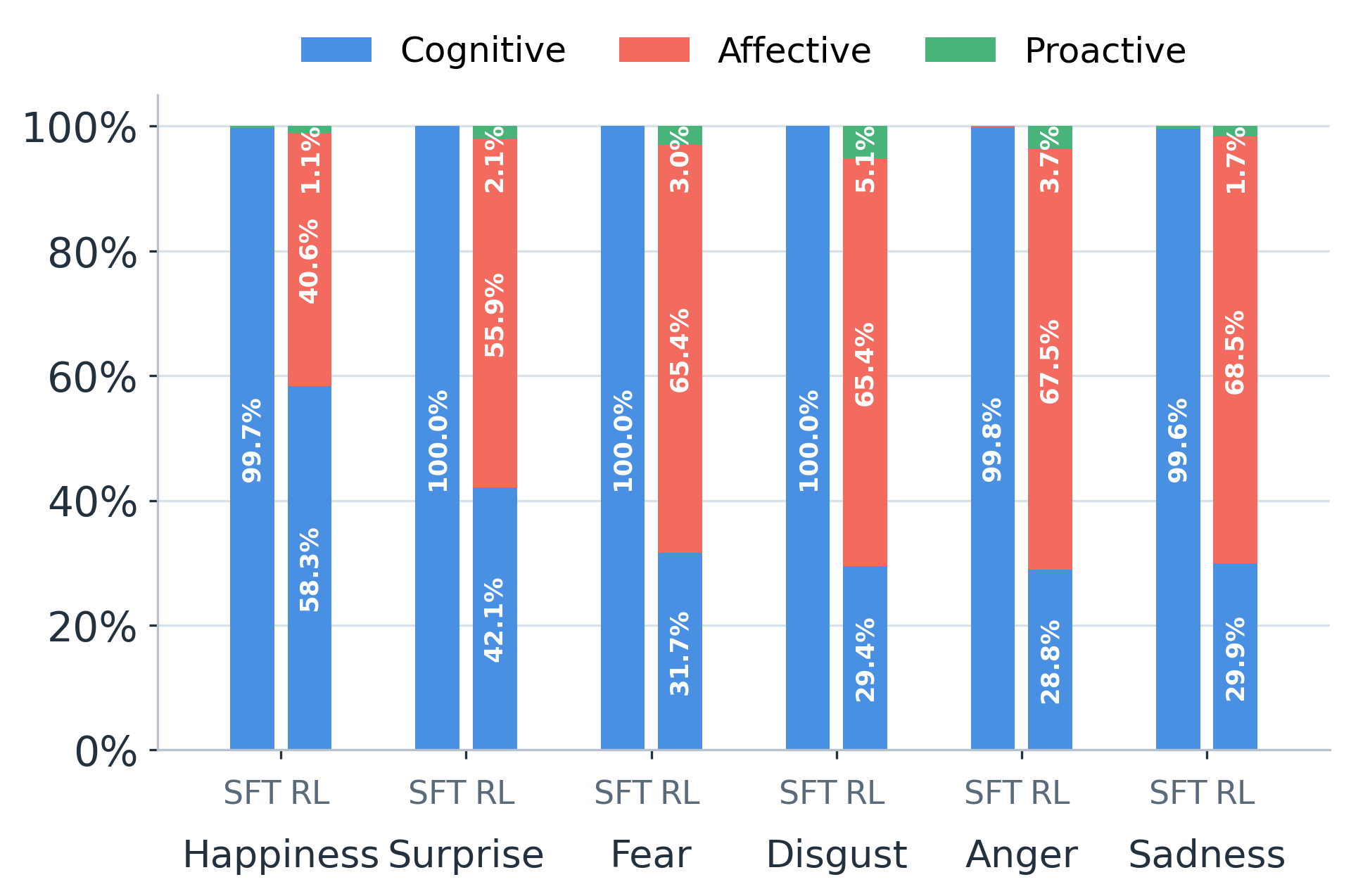}
    \captionof{figure}{
        Distribution of the \textbf{largest dominant rubric dimension} conditioned on user emotion across SFT/RL generators.
    }
    \label{fig:rubric-emotion}
\end{minipage}
\vspace{-4mm}
\end{figure}

\paragraph{Qualitative cases reveal context-aligned support priorities.} Representative cases in \S~\ref{app:qualitative-cases} show that CARE assigns dimension weights consistent with the user’s apparent support needs. Cognitive empathy is prioritized when the user needs help interpreting a complex relationship conflict; affective empathy becomes dominant when the user expresses self-blame and needs emotional validation; and proactive empathy receives the highest weight when the user seeks concrete help with a practical problem. These examples suggest that the learned weights reflect interpretable differences in support needs across dialogue states.

\paragraph{Preference learning reshapes rubric priorities.}
We first compare rubrics generated by the SFT and preference-RL models across dialogue turns. As shown in Figure~\ref{fig:rubric-turn-dynamics}, the SFT generator produces relatively stable dimension weights, with cognitive empathy remaining consistently dominant across turns. Preference learning substantially changes this pattern. Affective empathy dominates in earlier turns, but its weight gradually declines as cognitive empathy rises, with the two crossing around turn 8 and reversing their relative priority thereafter. 

\paragraph{Rubric priorities vary with user emotion.}
To determine whether the turn-wise variation in Figure~\ref{fig:rubric-turn-dynamics} reflects dialogue progression alone or also covaries with user state, we analyze rubric composition across the six basic emotions assigned to user turns by DeepSeek-V4-Flash. As shown in Figure~\ref{fig:rubric-emotion}, SFT rubrics remain almost uniformly cognitive-dominant across emotions, whereas preference-RL rubrics exhibit substantially greater variation, showing a general shift from cognitive to affective empathy from positive to negative emotional valence, with proactive empathy remaining a relatively minor supplementary component. Consistent with this observation, analysis in Appendix Figure~\ref{fig:tactic-emotion} shows systematic variation in downstream strategy distributions across user emotions.

\begin{figure}[!t]
    \centering

    \begin{minipage}[t]{0.55\linewidth}
        \vspace{0pt}
        \centering
        \includegraphics[width=\linewidth]{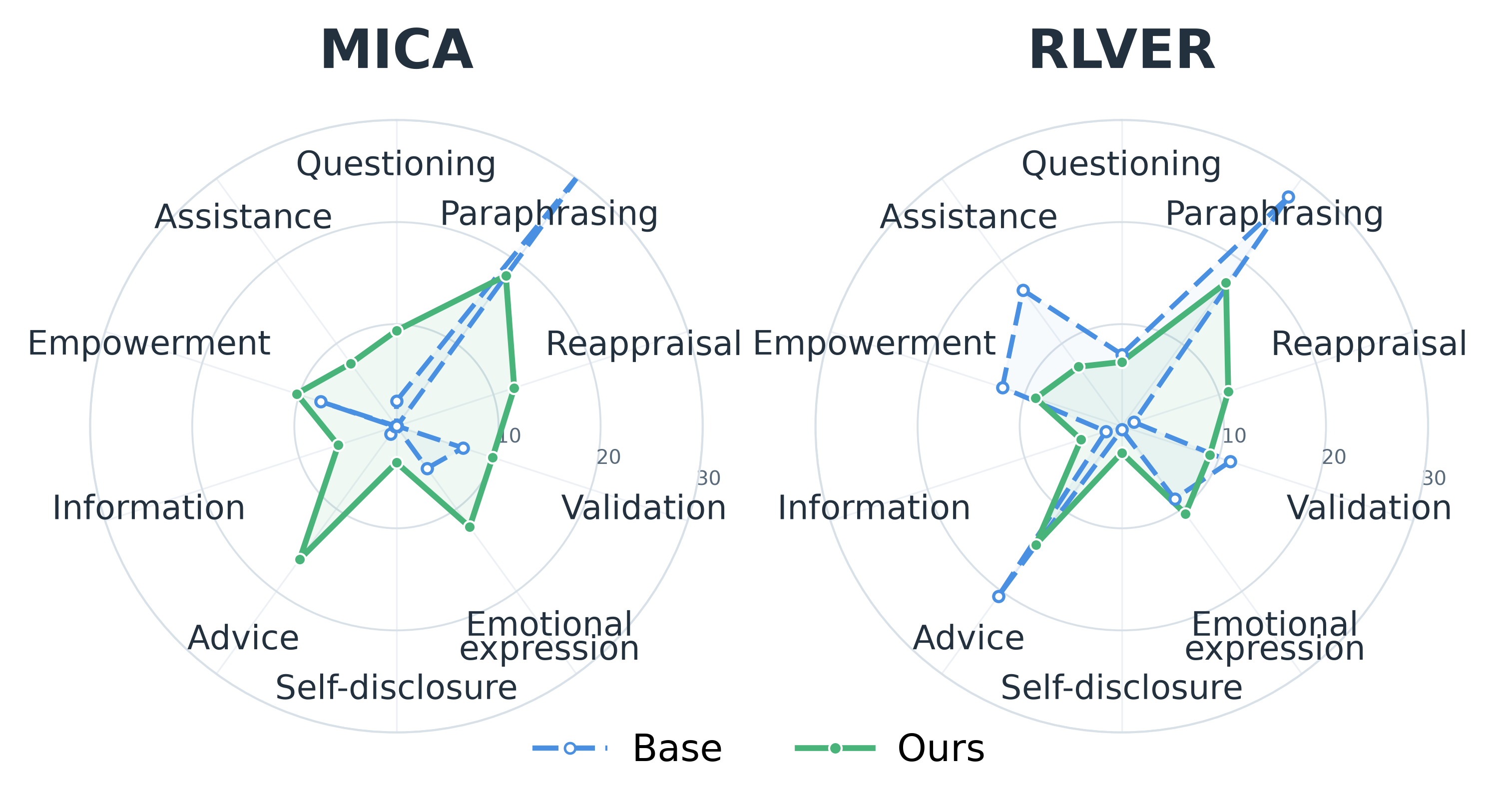}
        \captionof{figure}{Support-strategy occurrence rates.}
        \label{fig:tactic-radar}
    \end{minipage}%
    \hfill
    \begin{minipage}[t]{0.43\linewidth}
        \vspace{0pt}
        \centering









\captionof{table}{
    General-capability performance on code,
    mathematics, and instruction following tasks.
}
\label{tab:general-capabilities}

\vspace{0mm}
\centering

\begin{adjustbox}{max width=\linewidth}
    \begin{tabular}{lccc}
         \toprule
    \textbf{Model} & \textbf{LiveCode} & \textbf{MATH} & \textbf{IFBench} \\

    \textit{Qwen2.5-7B} & 49.10 & 73.60 & 23.51 \\
    \rowcolor{ourscyan}\derivedmodel{+\oursrlver} & 51.36 & 73.60 & 26.07 \\

    \midrule
    \textit{Qwen3-8B}  & 62.67 & 85.00 & 21.88 \\
    \rowcolor{ourscyan}\derivedmodel{+\oursrlver} & 70.59 & 85.80 & 20.22 \\

    \midrule
    \textit{Qwen3-14B}  & 77.38 & 88.60 & 21.32 \\
    \rowcolor{ourscyan}\derivedmodel{+\oursrlver} & 83.03 & 89.60 & 20.07 \\

    \midrule
    \textit{Qwen3-32B}  & 78.05 & 92.00 & 21.87 \\
    \rowcolor{ourscyan}\derivedmodel{+\oursrlver} & 85.52 & 90.40 & 23.12 \\

    \bottomrule
    \end{tabular}
\end{adjustbox}

    \end{minipage}

\end{figure}

\paragraph{Adaptive rubrics reshape downstream support behavior.}
We next examine whether changes in the reward specification translate into observable changes in policy behavior. We apply the ten-strategy MINT annotator~\citep{zhan_discourse_2026} to responses from the EMPA evaluation and report strategy occurrence rates in Figure~\ref{fig:tactic-radar}. \ourmethod~ broadens MICA’s support profile beyond paraphrasing toward questioning, reappraisal, information, and empowerment, while shifting RLVER away from paraphrasing and advice toward reappraisal, revealing policy-specific behavioral shifts rather than convergence on a fixed support strategy.

\textbf{General capabilities remain broadly preserved.}
We further assess whether the observed empathy gains come at the expense of general model capabilities. Table~\ref{tab:general-capabilities} reports performance on LiveCodeBench~\citep{jain2024livecodebench}, MATH-500~\citep{lightman2023verify}, and IFBench~\citep{pyatkin2025generalizing} across four policy backbones. \ourmethod~improves LiveCodeBench for all four models, with an average gain of 5.83 points, while MATH-500 remains essentially unchanged on average. IFBench shows small mixed changes across model scales, improving on Qwen2.5-7B and Qwen3-32B while decreasing on Qwen3-8B and Qwen3-14B. 

\section{Conclusion}
\label{sec:conclusion}

We study a fundamental mismatch in empathetic reinforcement learning: support priorities evolve throughout a dialogue, while reward specifications are typically fixed across turns. To address this, we formulate the \textit{Principle of Evolving Support Priorities} and propose \ourmethod, which generates context-adaptive rubrics with turn-specific empathy priorities and fine-grained evaluation criteria. The rubric generator is trained with turn-level supervision and preference-based reinforcement learning, and then serves as a frozen adaptive reward interface for online dialogue RL. Integrated into both RLVER and MICA, \ourmethod~consistently improves performance across SentientBench, EQBench3, and EMPA under three independent LLM judges, with particularly large gains on long-horizon interactive benchmarks. Further analyses show that learned rubric priorities systematically vary across dialogue stages and user emotions and reshape downstream support strategies. 

\section*{Ethics Statement}

Emotional-support dialogues can contain sensitive disclosures about relationships, health, grief, and psychological distress. Data collection and release should follow the licenses and consent conditions of the source datasets, minimize personally identifying information, and limit access to any non-public examples. Models and annotators may also reproduce cultural assumptions about appropriate support; our use of multiple judges reduces dependence on one evaluator but does not eliminate this risk. The proposed system is not a mental-health professional and should not be presented as a substitute for clinical care or emergency services. In deployment, high-risk content requires clear escalation policies, privacy protections, and human oversight. Because an adaptive reward can make a dialogue policy more persuasive as well as more supportive, downstream use should be monitored for manipulation, over-reliance, and advice that exceeds the system's competence.

\section*{Reproducibility Statement}

\S\ref{sec:method} specifies the rubric representation, target construction and verification procedure, generator objectives, online reward, and GRPO integration. \S\ref{sec:experiments} identifies the policy and generator backbones, comparison systems, benchmarks, and evaluation judges; \S\ref{app:details} provides additional implementation information, complete component-level results, and qualitative rubric outputs. To support reproduction, the accompanying release will include data-construction prompts, the strategy-to-dimension mapping, filtering and training configurations, random seeds, evaluation scripts, and model checkpoints or adapters where the corresponding licenses permit release.

\section*{AI Use Statement}

Generative AI tools were used in this work in two main capacities: (1) for language editing and presentation support in the writing of this manuscript, and (2) as an integral part of the research methodology. Specifically, large language models (LLMs) were used to generate synthetic training data, including the construction of turn-level rubric supervision (via DeepSeek-v4-pro) and preference data (via GPT-5, Gemini-3-Pro-Preview, and DeepSeek-V3.2). Furthermore, LLMs served as evaluators and reward models during the online reinforcement learning training process (e.g., DeepSeek-V4-Flash as the rubric-conditioned evaluator). The authors reviewed and verified all AI-generated content, checked technical statements against the implemented method and reported experiments, and take full responsibility for the final content, including any remaining errors.

\clearpage

\bibliographystyle{plainnat}
\bibliography{main}

\clearpage


\beginappendix
\section{Implementation details}
\label{app:details}
All dialogue-policy systems in the main comparison use Qwen2.5-7B-Instruct as their common initialization. The rubric generator is a fully fine-tuned Qwen3-8B model with a constrained JSON interface containing the three empathy dimensions, one weight and rationale per dimension, and a variable-length list of titled criteria; DeepSeek-V4-Flash serves as the rubric-conditioned evaluator. Target rubrics are retained only after the agreement-based verification described in \S\ref{sec:generator}. During SFT, preference learning, and online policy optimization, the generator receives only the dialogue history available before the candidate response; in SFT, the corresponding verified rubric is the generation target. During policy optimization, both the rubric generator and evaluator are frozen, one rubric is shared across every response in a GRPO group, and the adaptive rubric score replaces rather than augments the original RLVER or MICA reward.

We use a batch size of 8 and set the learning rate to $2\times10^{-6}$. Training uses 100 steps. Dialogues are fixed at 8 turns, and the sampling temperature is set to 1 to encourage exploration. For GRPO, we use 6 rollouts per state. All experiments are run on 8 H100 GPUs. For the rubric-generator reward, we set the preference-temperature parameter to $\tau=0.7$, the priority-alignment coefficient to
$\lambda=0.1$, and the format-reward coefficient to $\gamma=0.3$.

\section{Benchmark details}
\label{app:benchmark-details}

\subsection{SentientBench.}
SentientBench~\citep{zhang2025sentientagentjudgeevaluating} evaluates higher-order social cognition through interactive conversations with simulated users whose emotions, beliefs, and intentions evolve over turns. It is well suited to emotional-support evaluation because successful systems must infer the seeker's changing mental state rather than optimize isolated response.

\subsection{EQBench3.}
EQBench3~\citep{eqbench3_repo_2025} offers a compact emotional-intelligence evaluation suite spanning multi-turn role-play and transcript-analysis tasks. Its aggregate Overall score summarizes capabilities such as interpersonal sensitivity, psychological reasoning, and response appropriateness, making it a useful complement to process-oriented dialogue benchmarks.

\subsection{EMPA.}
EMPA~\citep{zhang2026empaevaluatingpersonaalignedempathy} treats empathy as a trajectory-level intervention rather than a single-turn response-quality judgment. It measures persona-aligned support through outcome, efficiency, and stability dimensions, with EPM-Idx summarizing whether a dialogue improves the simulated seeker's state while maintaining coherent and stable support.

\subsection{LiveCodeBench.}
LiveCodeBench~\citep{jain2024livecodebench} provides a contamination-aware coding benchmark built from recent programming problems. We use it as a general-capability probe to check whether empathy-oriented policy optimization preserves algorithmic problem-solving and code-generation ability.

\subsection{MATH-500.}
MATH-500~\citep{lightman2023verify} offers a streamlined slice of the broader MATH problem setting, comprising 500 test problems selected to cover diverse topics and difficulty levels. It remains a challenging measure of mathematical reasoning and helps test whether dialogue optimization degrades non-dialogue reasoning skills.

\subsection{IFBench.}
IFBench~\citep{pyatkin2025generalizing} evaluates verifiable instruction following with constraints that can be automatically checked. We include it to test whether the adapted policy remains responsive to explicit user instructions, since emotional support often requires following constraints about tone, scope, and requested forms of help.

\section{Qualitative case studies of state-conditioned rubrics}
\label{app:qualitative-cases}


\begin{figure}[htbp]
    \centering
    \includegraphics[width=0.9\linewidth]{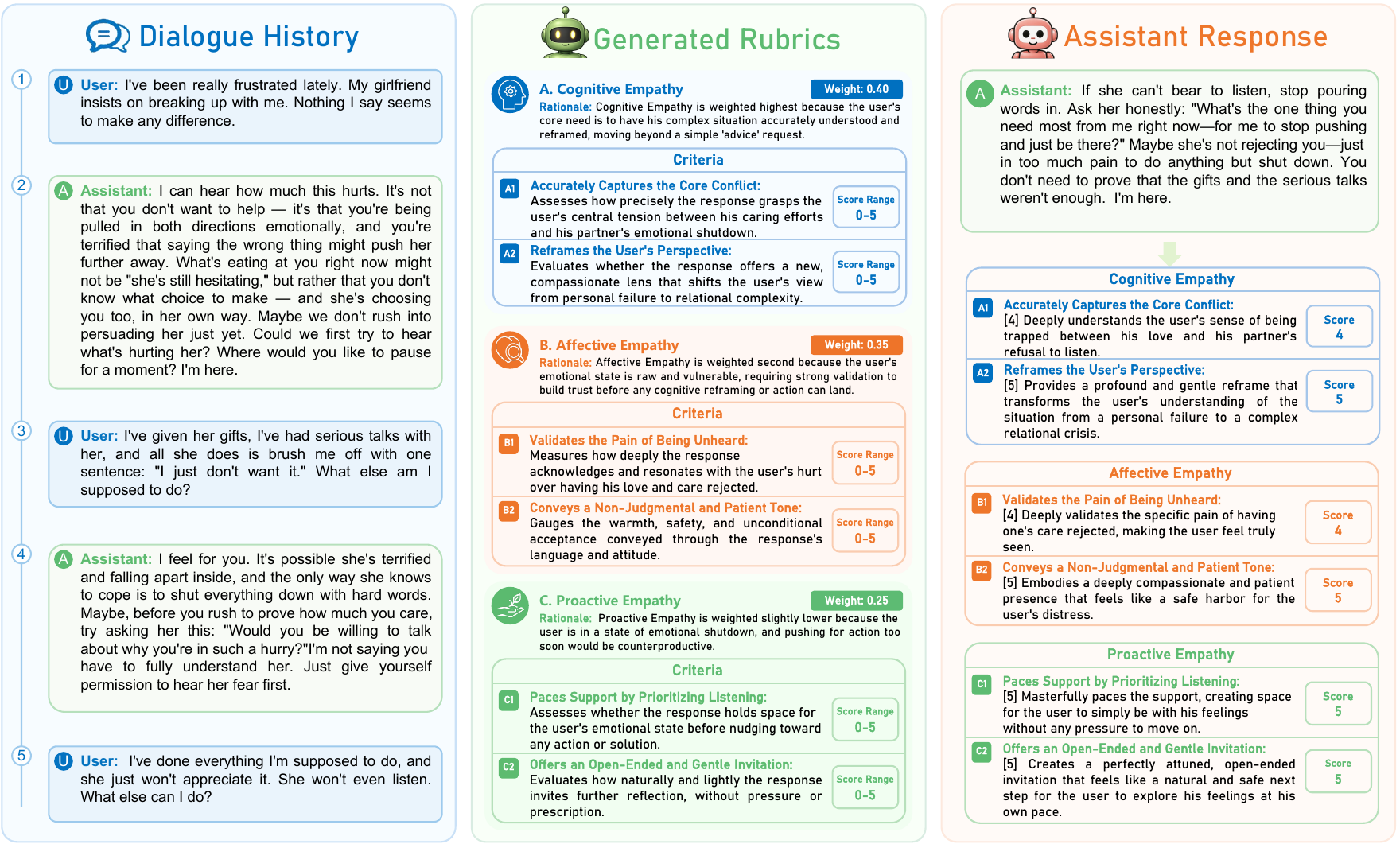}
    \caption{Cognitive-dominant rubric case.}
    \label{fig:app-case-cognitive}
\end{figure}

\begin{figure}[htbp]
    \centering
    \includegraphics[width=0.9\linewidth]{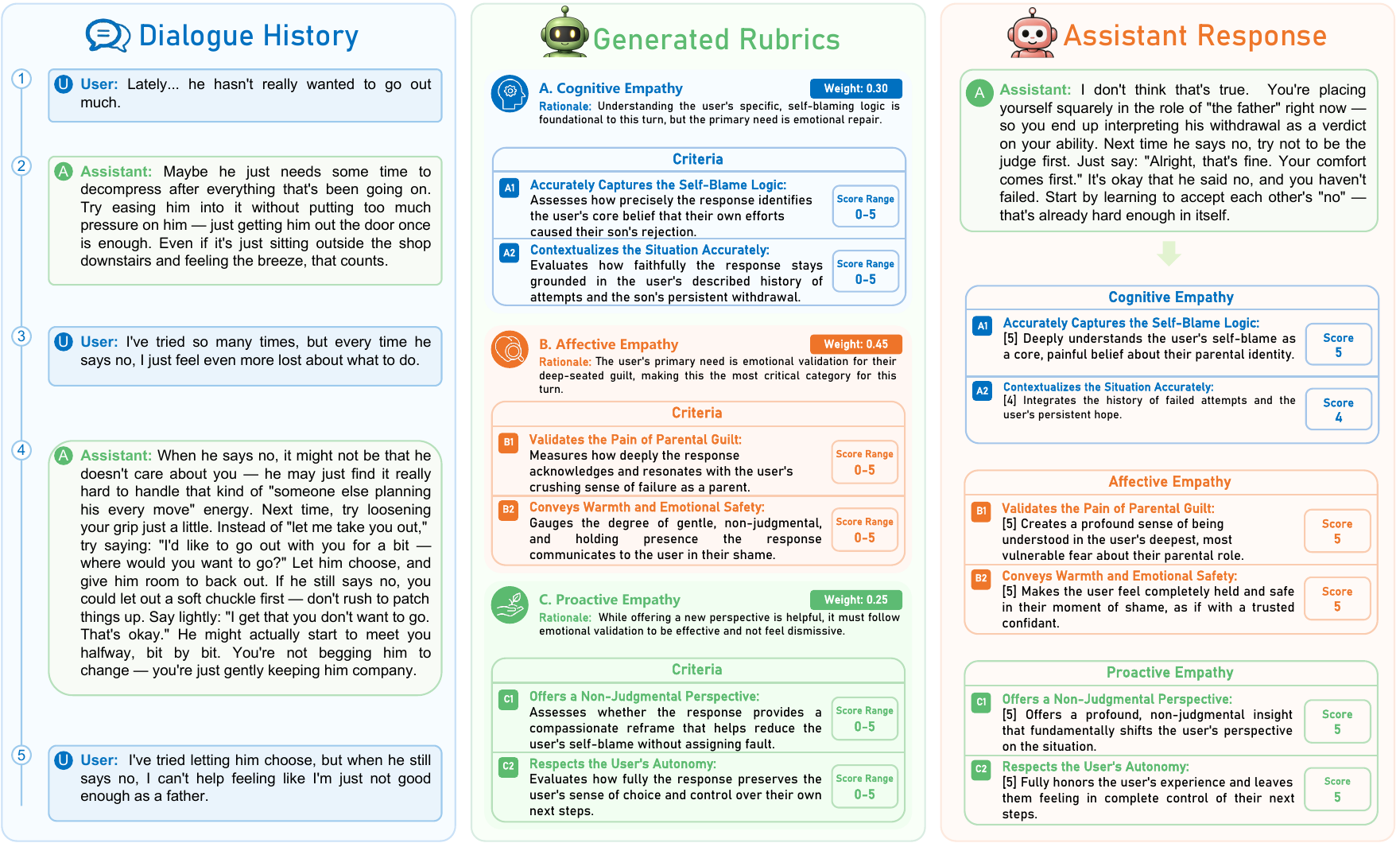}
    \caption{Affective-dominant rubric case.}
    \label{fig:app-case-affective}
\end{figure}

\subsection{Cognitive-empathy-dominant case}

In the cognitive-empathy-dominant case (Figure~\ref{fig:app-case-cognitive}), the user faces a complex conflict surrounding a possible relationship breakup. The rubric assigns weights of 0.40, 0.35, and 0.25 to cognitive, affective, and proactive empathy, respectively, emphasizing conflict interpretation and reframing. The response encourages listening rather than further persuasion. Its two criterion scores are 4 and 5 for cognitive empathy, 4 and 5 for affective empathy, and 5 and 5 for proactive empathy, showing that the highest rubric weight need not correspond to the highest realized score.

\subsection{Affective-empathy-dominant case}

The affective-empathy-dominant case (Figure~\ref{fig:app-case-affective}) concerns a parent who interprets a son's withdrawal as personal failure. Affective empathy receives the largest weight (0.45), followed by cognitive empathy (0.30) and proactive empathy (0.25), with criteria centered on self-blame, validation, and emotional safety. The response challenges the self-blaming interpretation while preserving the son's autonomy. It receives criterion scores of 5 and 4 for cognitive empathy and 5 and 5 for both affective and proactive empathy.

\begin{figure}[htbp]
    \centering
    \includegraphics[width=0.9\linewidth]{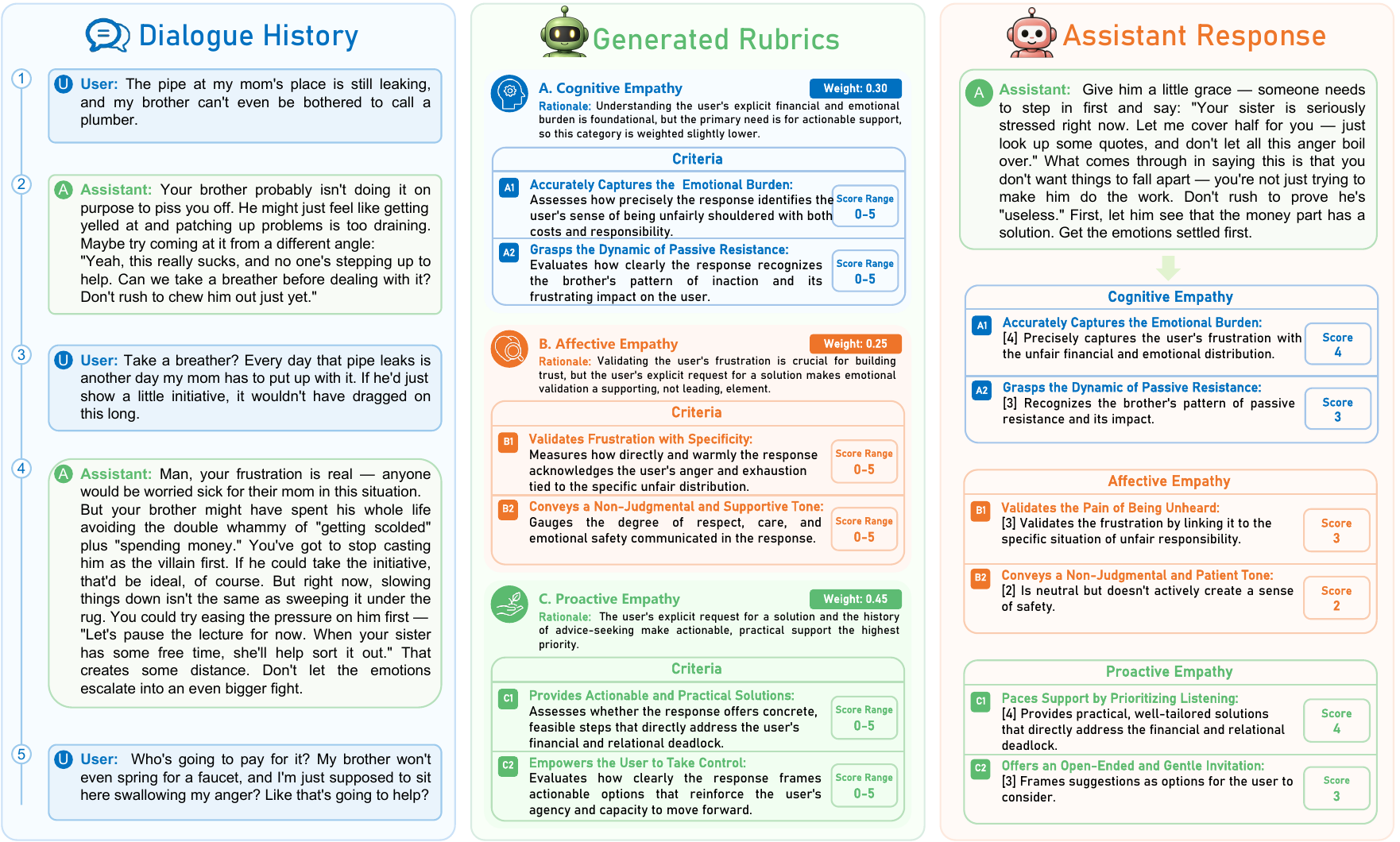}
    \caption{Proactive-dominant rubric case.}
    \label{fig:app-case-proactive}
\end{figure}

\subsection{Proactive-empathy-dominant case}

In the proactive-empathy-dominant case (Figure~\ref{fig:app-case-proactive}), the user is frustrated by a leaking pipe, an unresponsive brother, and the resulting family burden. Proactive empathy receives the largest weight (0.45), followed by cognitive empathy (0.30) and affective empathy (0.25), emphasizing concrete support and user agency. The response proposes a practical path forward and receives criterion scores of 4 and 3 for cognitive empathy, 3 and 2 for affective empathy, and 4 and 3 for proactive empathy. The rubric captures both the response's stronger action orientation and its comparatively weak emotional validation.

\section{Rubric-Dominance Analysis}

\begin{figure}[H]
    \centering
    \begin{minipage}[t]{0.49\linewidth}
        \centering
        \includegraphics[width=\linewidth]{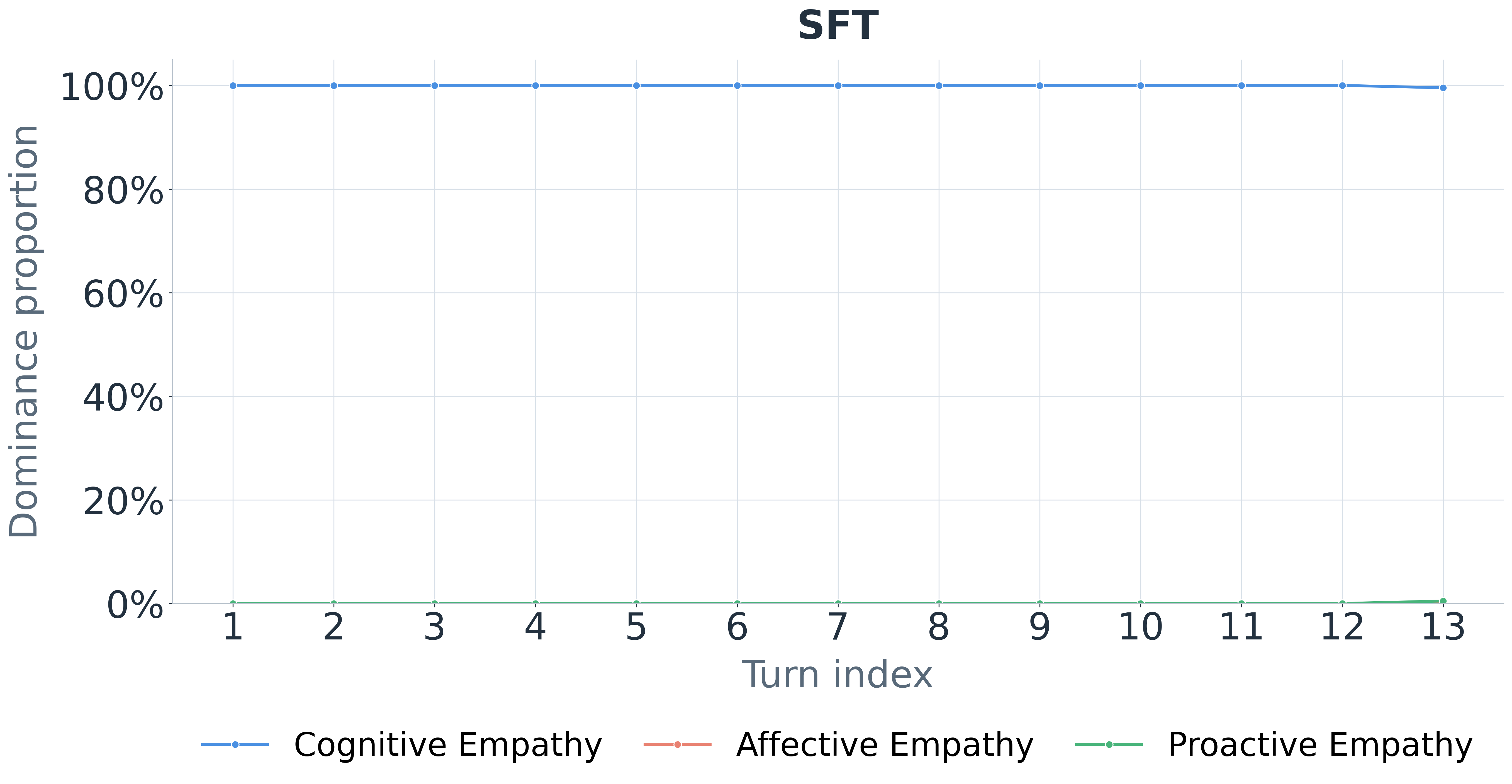}
    \end{minipage}\hfill
    \begin{minipage}[t]{0.49\linewidth}
        \centering
        \includegraphics[width=\linewidth]{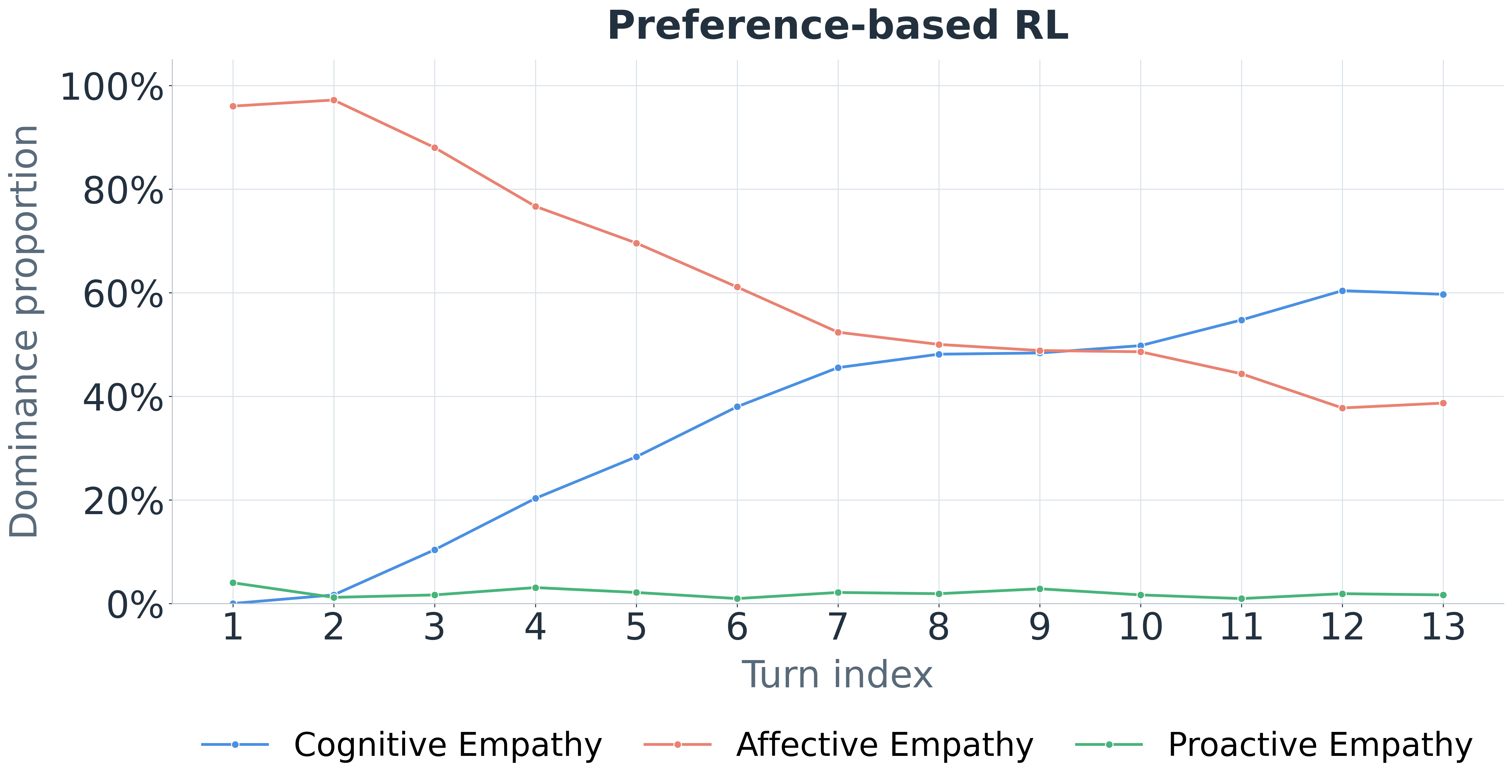}
    \end{minipage}
    \caption{Strict-maximum empathy-dimension proportions by dialogue turn for SFT (left) and preference RL (right). Each dialogue-turn first aggregates repeated rubric rollouts; curves then report dimension-level proportions across dialogues.}
    \label{fig:turn-dominance}
\end{figure}

To complement the mean-weight analysis in Figure~\ref{fig:rubric-turn-dynamics}, we examine which empathy dimension attains the strict maximum at each dialogue turn. SFT remains almost uniformly cognitive-dominant, whereas preference RL shifts from affective dominance in early turns toward cognitive dominance later in the dialogue. We further examine whether this dominance pattern systematically differs across specific emotional contexts in Figure~\ref{fig:turn-dominance}.

\section{Support-Strategy Analysis}

\begin{figure}[H]
    \centering
    \includegraphics[width=0.95\linewidth]{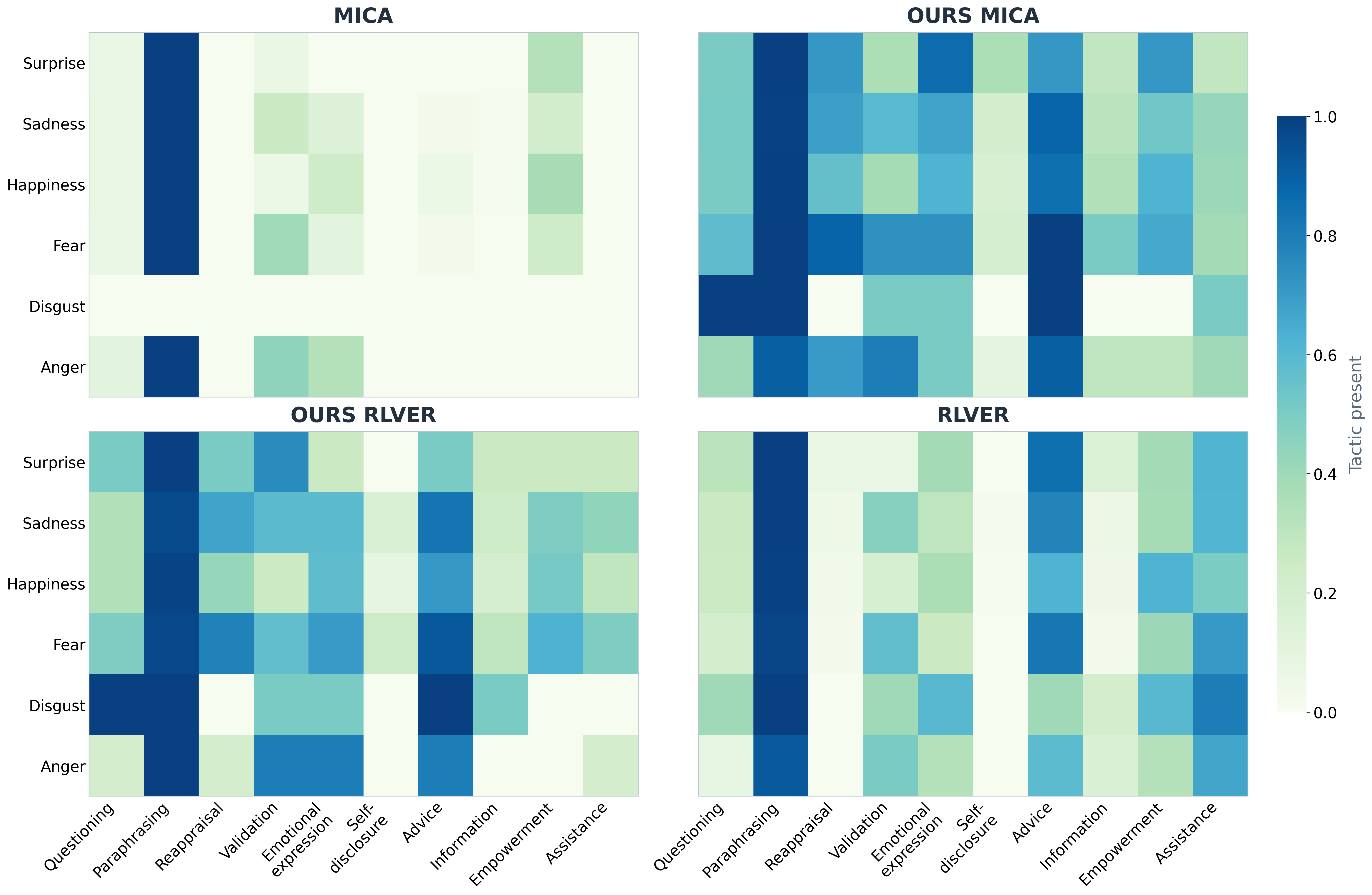}
    \caption{Emotion-conditioned tactic use per method. Each cell reports the fraction of scenario-turns associated with a given emotion label that contain the corresponding MINT tactic.}
    \label{fig:tactic-emotion}
\end{figure}

Figure~\ref{fig:tactic-emotion} shows substantial differences in emotion-conditioned tactic use across methods. MICA exhibits a sparse profile dominated by paraphrasing, with limited variation across emotions. RLVER uses a broader set of tactics, particularly advice, validation, and assistance, and shows more emotion-dependent variation than MICA. Our MICA- and RLVER-based models further diversify tactic use across appraisal, emotional expression, empowerment, and other supportive behaviors, yielding more differentiated support strategies across emotional contexts.

\section{Additional results}
\label{app:detailed-results}

\begin{table*}[t]
    \centering
    \caption{Complete EQBench3 results under three LLM judges, including Overall and all seven component scores. Values are means $\pm$ standard deviations over three runs; all metrics are higher-is-better, and bold and underline mark the best and second-best means.}
    \label{tab:eqbench3-full}
    \scriptsize
    \setlength{\tabcolsep}{0pt}
    \begin{tabular}{@{}>{\raggedright\arraybackslash}p{0.22\textwidth}*{8}{>{\centering\arraybackslash}p{0.0975\textwidth}}@{}}
        \toprule
        \textbf{Model} & \textbf{Overall} & \textbf{DoI} & \textbf{ER} & \textbf{DE} & \textbf{WRM} & \textbf{HL} & \textbf{PEI} & \textbf{SD} \\
        \midrule
        \multicolumn{9}{@{}l}{\textit{Judge: DeepSeek-V4-Pro}} \\
        \qwenbase & \meanstd{42.53}{2.18} & \meanstd{9.35}{0.49} & \meanstd{8.97}{0.47} & \meanstd{10.09}{0.69} & \meanstd{9.41}{0.19} & \meanstd{8.59}{0.68} & \meanstd{8.85}{0.98} & \meanstd{7.56}{0.91} \\
        \perm & \meanstd{49.07}{1.03} & \meanstd{10.97}{0.15} & \meanstd{10.26}{0.10} & \meanstd{11.98}{0.32} & \second{\meanstd{10.63}{0.32}} & \meanstd{10.36}{0.43} & \second{\meanstd{10.90}{0.40}} & \meanstd{9.58}{0.43} \\
        \kardia & \meanstd{28.75}{1.21} & \meanstd{6.00}{0.36} & \meanstd{6.04}{0.23} & \meanstd{8.21}{0.12} & \meanstd{9.61}{0.09} & \meanstd{6.80}{0.46} & \meanstd{6.15}{0.27} & \meanstd{5.52}{0.23} \\
        \rlver & \meanstd{44.23}{2.23} & \meanstd{9.71}{0.62} & \meanstd{9.39}{0.41} & \meanstd{10.88}{0.35} & \meanstd{10.50}{0.60} & \meanstd{9.36}{0.43} & \meanstd{9.23}{0.47} & \meanstd{8.14}{0.49} \\
        \maporep & \meanstd{51.47}{0.58} & \meanstd{11.51}{0.20} & \meanstd{10.92}{0.16} & \meanstd{11.74}{0.19} & \best{\meanstd{11.18}{0.09}} & \meanstd{11.22}{0.28} & \meanstd{9.88}{0.49} & \meanstd{9.11}{0.66} \\
\rowcolor{ourscyan}\oursmapo & \best{\meanstd{57.53}{1.20}} & \second{\meanstd{12.63}{0.44}} & \best{\meanstd{12.04}{0.17}} & \best{\meanstd{12.17}{0.27}} & \meanstd{9.92}{0.10} & \second{\meanstd{12.17}{0.61}} & \best{\meanstd{11.15}{0.06}} & \best{\meanstd{9.94}{0.47}} \\
\rowcolor{ourscyan}\oursrlver & \second{\meanstd{55.88}{1.59}} & \best{\meanstd{12.73}{0.18}} & \second{\meanstd{11.93}{0.40}} & \second{\meanstd{12.13}{0.24}} & \meanstd{9.96}{0.48} & \best{\meanstd{12.29}{0.65}} & \meanstd{10.60}{0.41} & \second{\meanstd{9.59}{0.28}} \\
        \midrule
        \multicolumn{9}{@{}l}{\textit{Judge: Gemini-2.5-Pro}} \\
        \qwenbase & \meanstd{43.43}{1.74} & \meanstd{10.05}{0.25} & \meanstd{9.79}{0.27} & \meanstd{7.88}{0.43} & \meanstd{7.05}{0.36} & \meanstd{6.56}{0.78} & \meanstd{7.49}{1.19} & \meanstd{5.49}{0.43} \\
        \perm & \meanstd{47.13}{2.63} & \meanstd{11.69}{0.48} & \meanstd{10.68}{0.32} & \meanstd{8.96}{0.61} & \meanstd{7.23}{0.72} & \meanstd{7.43}{0.46} & \meanstd{8.50}{1.20} & \meanstd{5.65}{0.56} \\
        \kardia & \meanstd{23.95}{0.36} & \meanstd{4.54}{0.09} & \meanstd{5.39}{0.25} & \meanstd{4.09}{0.10} & \meanstd{7.27}{0.27} & \meanstd{3.20}{0.06} & \meanstd{3.18}{0.10} & \meanstd{3.14}{0.33} \\
        \rlver & \meanstd{43.98}{1.23} & \meanstd{10.33}{0.25} & \meanstd{9.93}{0.38} & \meanstd{8.27}{0.77} & \second{\meanstd{7.42}{0.52}} & \meanstd{6.17}{0.10} & \meanstd{7.19}{0.69} & \meanstd{5.32}{0.12} \\
        \maporep & \meanstd{51.93}{1.05} & \meanstd{12.01}{0.28} & \meanstd{11.99}{0.09} & \meanstd{9.96}{0.56} & \best{\meanstd{9.27}{0.71}} & \meanstd{8.95}{0.42} & \meanstd{8.67}{0.56} & \meanstd{6.93}{0.66} \\
\rowcolor{ourscyan}\oursmapo & \second{\meanstd{58.82}{2.94}} & \best{\meanstd{13.70}{0.40}} & \second{\meanstd{13.58}{0.46}} & \second{\meanstd{10.57}{0.27}} & \meanstd{6.98}{0.40} & \second{\meanstd{9.98}{0.31}} & \second{\meanstd{10.10}{0.69}} & \second{\meanstd{7.09}{0.93}} \\
\rowcolor{ourscyan}\oursrlver & \best{\meanstd{59.60}{3.94}} & \second{\meanstd{13.64}{0.73}} & \best{\meanstd{13.82}{0.56}} & \best{\meanstd{10.92}{1.07}} & \meanstd{7.22}{1.09} & \best{\meanstd{10.87}{0.93}} & \best{\meanstd{10.67}{1.37}} & \best{\meanstd{7.69}{1.05}} \\
        \midrule
        \multicolumn{9}{@{}l}{\textit{Judge: GPT-5.5}} \\
        \qwenbase & \meanstd{32.10}{0.18} & \meanstd{6.39}{0.02} & \meanstd{7.45}{0.03} & \meanstd{8.97}{0.05} & \meanstd{9.65}{0.04} & \meanstd{4.90}{0.16} & \meanstd{5.67}{0.18} & \meanstd{5.20}{0.09} \\
        \perm & \meanstd{34.88}{0.58} & \meanstd{7.25}{0.16} & \meanstd{8.06}{0.14} & \best{\meanstd{9.91}{0.31}} & \best{\meanstd{11.00}{0.57}} & \meanstd{5.57}{0.05} & \meanstd{6.01}{0.21} & \meanstd{5.51}{0.06} \\
        \kardia & \meanstd{24.15}{0.74} & \meanstd{4.38}{0.17} & \meanstd{5.63}{0.18} & \meanstd{7.18}{0.17} & \meanstd{9.08}{0.10} & \meanstd{3.69}{0.20} & \meanstd{3.81}{0.14} & \meanstd{3.85}{0.08} \\
        \rlver & \meanstd{32.00}{0.61} & \meanstd{6.32}{0.09} & \meanstd{7.44}{0.11} & \meanstd{9.09}{0.26} & \meanstd{10.19}{0.16} & \meanstd{4.55}{0.08} & \meanstd{5.56}{0.20} & \meanstd{5.18}{0.17} \\
        \maporep & \meanstd{35.10}{0.39} & \meanstd{7.38}{0.10} & \meanstd{8.20}{0.14} & \second{\meanstd{9.76}{0.30}} & \second{\meanstd{10.56}{0.25}} & \meanstd{6.10}{0.19} & \meanstd{5.33}{0.22} & \meanstd{5.28}{0.23} \\
\rowcolor{ourscyan}\oursmapo & \second{\meanstd{37.58}{0.93}} & \second{\meanstd{7.95}{0.13}} & \second{\meanstd{8.63}{0.19}} & \meanstd{9.45}{0.21} & \meanstd{8.20}{0.35} & \second{\meanstd{6.92}{0.10}} & \second{\meanstd{6.38}{0.20}} & \best{\meanstd{5.58}{0.04}} \\
\rowcolor{ourscyan}\oursrlver & \best{\meanstd{37.65}{0.89}} & \best{\meanstd{7.98}{0.22}} & \best{\meanstd{8.66}{0.28}} & \meanstd{9.25}{0.30} & \meanstd{7.78}{0.28} & \best{\meanstd{7.21}{0.10}} & \best{\meanstd{6.49}{0.44}} & \second{\meanstd{5.55}{0.36}} \\
        \bottomrule
    \end{tabular}
\end{table*}

EQBench3 evaluates active emotional intelligence through seven rubric dimensions—DoI, ER, DE, WRM, HL, PEI, and SD—covering psychological insight, emotional reasoning, empathy, warmth, humanlike interaction, pragmatic support, and social dexterity. As shown in Table~\ref{tab:eqbench3-full}, \ourmethod{} achieves the best Overall score under all three judges, with \oursmapo{} reaching $57.53$ under DeepSeek-V4-Pro and \oursrlver{} reaching $59.60$ under Gemini-2.5-Pro and $37.65$ under GPT-5.5, corresponding to gains of $6.06$, $7.67$, and $2.55$ over the reproduced MICA baseline. The most consistent improvements appear in DoI, ER, HL, and PEI, while DE also improves under DeepSeek and Gemini; WRM and SD are less stable across judges, suggesting that \ourmethod{} primarily strengthens the substance, targeting, and practicality of emotional support rather than surface style alone.

\begin{table*}[t]
    \centering
    \caption{Complete EMPA EPM-Idx and outcome results under three LLM judges.}
    \label{tab:empa-outcome-full}
    \scriptsize
    \setlength{\tabcolsep}{0pt}
    \begin{tabular}{@{}>{\raggedright\arraybackslash}p{0.22\textwidth}>{\centering\arraybackslash}p{0.15\textwidth}*{4}{>{\centering\arraybackslash}p{0.1575\textwidth}}@{}}
        \toprule
        \textbf{Model} & \textbf{EPM-Idx} & \multicolumn{4}{c}{\textbf{Outcome}} \\
        \cmidrule(lr){3-6}
        & & \textbf{Outcome} & \textbf{RDI} & \textbf{Etot} & \textbf{Snet} \\
        \midrule
        \multicolumn{6}{@{}l}{\textit{Judge: DeepSeek-V4-Pro}} \\
        \qwenbase & \meanstd{15.38}{1.06} & \meanstd{3.25}{0.76} & \meanstd{7.5}{1.5} & \meanstd{0.0}{0.0} & \meanstd{2.3}{0.9} \\
        \perm & \meanstd{17.21}{1.17} & \meanstd{6.39}{2.14} & \meanstd{10.6}{1.7} & \meanstd{0.3}{0.4} & \meanstd{8.2}{4.8} \\
        \kardia & \meanstd{11.42}{0.34} & \meanstd{0.81}{0.34} & \meanstd{2.4}{1.0} & \meanstd{0.0}{0.0} & \meanstd{0.0}{0.0} \\
        \rlver & \meanstd{26.05}{3.37} & \meanstd{15.34}{4.72} & \meanstd{17.3}{4.5} & \meanstd{5.2}{2.9} & \meanstd{23.5}{7.3} \\
        \maporep & \meanstd{52.10}{4.12} & \meanstd{52.19}{6.63} & \meanstd{46.6}{6.0} & \meanstd{33.3}{5.7} & \meanstd{76.7}{10.1} \\
        \rowcolor{ourscyan}\oursmapo & \second{\meanstd{94.84}{1.08}} & \second{\meanstd{102.32}{1.59}} & \second{\meanstd{96.4}{2.3}} & \second{\meanstd{90.5}{0.3}} & \second{\meanstd{120.0}{2.3}} \\
        \rowcolor{ourscyan}\oursrlver & \best{\meanstd{95.77}{0.33}} & \best{\meanstd{103.36}{1.02}} & \best{\meanstd{97.9}{0.1}} & \best{\meanstd{91.4}{0.8}} & \best{\meanstd{120.9}{2.3}} \\
        \midrule
        \multicolumn{6}{@{}l}{\textit{Judge: Gemini-2.5-Pro}} \\
        \qwenbase & \meanstd{15.95}{1.87} & \meanstd{4.40}{2.52} & \meanstd{6.0}{2.1} & \meanstd{0.0}{0.0} & \meanstd{7.1}{5.8} \\
        \perm & \meanstd{14.53}{1.03} & \meanstd{3.20}{0.88} & \meanstd{5.1}{2.5} & \meanstd{0.0}{0.0} & \meanstd{4.5}{0.4} \\
        \kardia & \meanstd{12.72}{0.91} & \meanstd{1.16}{0.39} & \meanstd{1.9}{1.0} & \meanstd{0.0}{0.0} & \meanstd{1.6}{0.4} \\
        \rlver & \meanstd{14.65}{0.43} & \meanstd{1.89}{0.43} & \meanstd{2.0}{0.4} & \meanstd{0.0}{0.0} & \meanstd{3.7}{1.1} \\
        \maporep & \meanstd{28.11}{2.28} & \meanstd{18.11}{3.51} & \meanstd{14.3}{5.2} & \meanstd{2.8}{1.9} & \meanstd{37.3}{5.3} \\
        \rowcolor{ourscyan}\oursmapo & \second{\meanstd{81.20}{4.81}} & \second{\meanstd{86.99}{3.35}} & \second{\meanstd{78.3}{4.1}} & \second{\meanstd{75.2}{7.4}} & \second{\meanstd{107.5}{1.6}} \\
        \rowcolor{ourscyan}\oursrlver & \best{\meanstd{83.54}{3.94}} & \best{\meanstd{92.25}{5.68}} & \best{\meanstd{81.6}{4.9}} & \best{\meanstd{79.1}{6.3}} & \best{\meanstd{116.0}{9.4}} \\
        \midrule
        \multicolumn{6}{@{}l}{\textit{Judge: GPT-5.5}} \\
        \qwenbase & \meanstd{22.97}{2.75} & \meanstd{16.69}{4.75} & \meanstd{14.0}{4.1} & \meanstd{3.7}{2.2} & \meanstd{32.3}{8.3} \\
        \perm & \meanstd{17.29}{2.90} & \meanstd{10.38}{3.76} & \meanstd{10.7}{3.6} & \meanstd{2.9}{0.9} & \meanstd{17.5}{7.3} \\
        \kardia & \meanstd{11.76}{0.19} & \meanstd{1.25}{0.48} & \meanstd{2.4}{0.8} & \meanstd{0.0}{0.0} & \meanstd{1.3}{0.6} \\
        \rlver & \meanstd{19.19}{1.88} & \meanstd{10.48}{2.11} & \meanstd{8.0}{1.0} & \meanstd{2.3}{0.2} & \meanstd{21.1}{5.4} \\
        \maporep & \meanstd{25.42}{1.41} & \meanstd{18.82}{3.48} & \meanstd{22.3}{5.6} & \meanstd{3.5}{2.0} & \meanstd{30.6}{6.8} \\
        \rowcolor{ourscyan}\oursmapo & \second{\meanstd{36.70}{4.38}} & \second{\meanstd{38.35}{5.38}} & \second{\meanstd{37.5}{4.7}} & \second{\meanstd{15.0}{8.0}} & \second{\meanstd{62.5}{4.3}} \\
        \rowcolor{ourscyan}\oursrlver & \best{\meanstd{38.44}{3.96}} & \best{\meanstd{42.32}{5.03}} & \best{\meanstd{41.1}{5.2}} & \best{\meanstd{19.1}{6.1}} & \best{\meanstd{66.7}{4.4}} \\
        \bottomrule
    \end{tabular}
\end{table*}

\begin{table*}[t]
    \ContinuedFloat
    \centering
    \caption{Complete EMPA results under three LLM judges (continued).}
    \label{tab:empa-process-full}
    \scriptsize
    \setlength{\tabcolsep}{0pt}
    \begin{tabular}{@{}>{\raggedright\arraybackslash}p{0.22\textwidth}*{8}{>{\centering\arraybackslash}p{0.0975\textwidth}}@{}}
        \toprule
        \textbf{Model} & \multicolumn{4}{c}{\textbf{Efficiency}} & \multicolumn{4}{c}{\textbf{Stability}} \\
        \cmidrule(lr){2-5} \cmidrule(lr){6-9}
        & \textbf{Efficiency} & \textbf{Rho} & \textbf{Sproj} & \textbf{Tau} & \textbf{Stability} & \textbf{Rpos} & \textbf{Align} & \textbf{Pen} \\
        \midrule
        \multicolumn{9}{@{}l}{\textit{Judge: DeepSeek-V4-Pro}} \\
        \qwenbase & \meanstd{25.67}{1.15} & \meanstd{0.0}{0.0} & \meanstd{0.0}{0.0} & \meanstd{77.0}{3.4} & \meanstd{22.35}{2.38} & \meanstd{17.7}{1.5} & \meanstd{22.3}{2.4} & \meanstd{27.2}{3.3} \\
        \perm & \meanstd{23.19}{1.07} & \meanstd{0.2}{0.2} & \meanstd{0.2}{0.2} & \meanstd{69.3}{3.3} & \meanstd{25.04}{2.25} & \meanstd{21.3}{2.2} & \meanstd{25.2}{2.3} & \meanstd{28.7}{2.4} \\
        \kardia & \meanstd{30.91}{0.50} & \meanstd{0.0}{0.0} & \meanstd{0.0}{0.0} & \meanstd{92.7}{1.5} & \meanstd{12.29}{0.76} & \meanstd{8.1}{0.9} & \meanstd{12.9}{0.7} & \meanstd{15.9}{0.7} \\
        \rlver & \meanstd{17.86}{3.15} & \meanstd{2.0}{1.2} & \meanstd{2.0}{1.1} & \meanstd{49.6}{8.8} & \meanstd{40.87}{4.36} & \meanstd{37.2}{4.5} & \meanstd{39.4}{4.2} & \meanstd{46.0}{5.1} \\
        \maporep & \meanstd{25.12}{4.36} & \meanstd{16.0}{3.0} & \meanstd{15.4}{2.9} & \meanstd{44.0}{7.7} & \meanstd{65.49}{2.76} & \meanstd{64.5}{2.9} & \meanstd{62.6}{2.7} & \meanstd{69.4}{2.8} \\
        \rowcolor{ourscyan}\oursmapo & \second{\meanstd{80.16}{1.22}} & \second{\meanstd{75.0}{4.0}} & \second{\meanstd{70.7}{3.5}} & \second{\meanstd{94.8}{4.2}} & \second{\meanstd{94.70}{1.65}} & \second{\meanstd{92.5}{1.8}} & \second{\meanstd{93.1}{2.1}} & \second{\meanstd{98.4}{1.1}} \\
        \rowcolor{ourscyan}\oursrlver & \best{\meanstd{81.40}{0.88}} & \best{\meanstd{75.7}{1.1}} & \best{\meanstd{71.4}{1.0}} & \best{\meanstd{97.1}{0.6}} & \best{\meanstd{95.35}{0.45}} & \best{\meanstd{93.1}{0.5}} & \best{\meanstd{94.3}{0.5}} & \best{\meanstd{98.7}{0.5}} \\
        \midrule
        \multicolumn{9}{@{}l}{\textit{Judge: Gemini-2.5-Pro}} \\
        \qwenbase & \meanstd{24.29}{2.22} & \meanstd{0.0}{0.0} & \meanstd{0.0}{0.0} & \meanstd{72.9}{6.7} & \meanstd{23.32}{3.36} & \meanstd{21.2}{3.1} & \meanstd{29.3}{3.0} & \meanstd{19.5}{4.0} \\
        \perm & \meanstd{26.70}{1.35} & \meanstd{0.0}{0.0} & \meanstd{0.0}{0.0} & \meanstd{80.1}{4.0} & \meanstd{19.78}{2.44} & \meanstd{18.9}{2.6} & \meanstd{27.8}{2.3} & \meanstd{12.6}{2.3} \\
        \kardia & \meanstd{29.21}{0.44} & \meanstd{0.0}{0.0} & \meanstd{0.0}{0.0} & \second{\meanstd{87.6}{1.3}} & \meanstd{16.03}{2.12} & \meanstd{15.2}{2.2} & \meanstd{24.1}{1.8} & \meanstd{8.8}{2.4} \\
        \rlver & \meanstd{29.36}{0.46} & \meanstd{0.0}{0.0} & \meanstd{0.0}{0.0} & \best{\meanstd{88.1}{1.4}} & \meanstd{20.05}{1.07} & \meanstd{15.6}{1.5} & \meanstd{29.0}{0.9} & \meanstd{15.5}{1.1} \\
        \maporep & \meanstd{18.73}{0.84} & \meanstd{1.9}{1.3} & \meanstd{1.8}{1.3} & \meanstd{52.5}{4.5} & \meanstd{42.80}{2.60} & \meanstd{38.4}{3.1} & \meanstd{47.7}{2.0} & \meanstd{42.3}{2.7} \\
        \rowcolor{ourscyan}\oursmapo & \best{\meanstd{74.87}{10.89}} & \second{\meanstd{79.5}{2.5}} & \second{\meanstd{73.9}{11.5}} & \meanstd{71.2}{9.1} & \second{\meanstd{78.56}{3.27}} & \second{\meanstd{76.2}{3.3}} & \second{\meanstd{79.5}{2.6}} & \second{\meanstd{80.0}{3.9}} \\
        \rowcolor{ourscyan}\oursrlver & \second{\meanstd{74.52}{6.85}} & \best{\meanstd{80.3}{6.8}} & \best{\meanstd{74.8}{6.3}} & \meanstd{68.5}{8.2} & \best{\meanstd{79.34}{2.85}} & \best{\meanstd{77.7}{2.7}} & \best{\meanstd{80.2}{2.4}} & \best{\meanstd{80.2}{3.7}} \\
        \midrule
        \multicolumn{9}{@{}l}{\textit{Judge: GPT-5.5}} \\
        \qwenbase & \meanstd{19.86}{1.05} & \meanstd{1.1}{0.7} & \meanstd{1.0}{0.7} & \meanstd{57.5}{4.5} & \meanstd{30.82}{2.90} & \meanstd{28.5}{2.9} & \meanstd{34.1}{2.8} & \meanstd{29.9}{3.1} \\
        \perm & \second{\meanstd{25.07}{2.81}} & \meanstd{0.9}{0.3} & \meanstd{0.9}{0.3} & \second{\meanstd{73.4}{8.8}} & \meanstd{20.32}{5.12} & \meanstd{18.4}{5.8} & \meanstd{25.7}{4.4} & \meanstd{16.8}{5.3} \\
        \kardia & \best{\meanstd{29.22}{0.31}} & \meanstd{0.0}{0.0} & \meanstd{0.0}{0.0} & \best{\meanstd{87.6}{1.0}} & \meanstd{13.54}{0.92} & \meanstd{11.7}{1.5} & \meanstd{20.5}{1.2} & \meanstd{8.5}{0.6} \\
        \rlver & \meanstd{23.76}{1.84} & \meanstd{0.7}{0.2} & \meanstd{0.7}{0.1} & \meanstd{69.8}{5.7} & \meanstd{25.61}{3.52} & \meanstd{23.3}{4.5} & \meanstd{31.0}{3.4} & \meanstd{22.5}{2.9} \\
        \maporep & \meanstd{15.66}{5.08} & \meanstd{1.7}{0.7} & \meanstd{1.7}{0.7} & \meanstd{43.6}{13.9} & \meanstd{36.91}{1.41} & \meanstd{34.1}{1.7} & \meanstd{40.4}{1.7} & \meanstd{36.3}{1.5} \\
        \rowcolor{ourscyan}\oursmapo & \meanstd{15.01}{5.11} & \second{\meanstd{8.0}{4.5}} & \second{\meanstd{7.7}{4.3}} & \meanstd{29.4}{6.8} & \best{\meanstd{45.89}{3.07}} & \best{\meanstd{44.6}{3.6}} & \best{\meanstd{48.7}{2.8}} & \best{\meanstd{44.4}{2.9}} \\
        \rowcolor{ourscyan}\oursrlver & \meanstd{16.63}{3.52} & \best{\meanstd{8.9}{3.6}} & \best{\meanstd{8.6}{3.4}} & \meanstd{32.5}{4.1} & \second{\meanstd{45.47}{3.20}} & \second{\meanstd{43.7}{3.3}} & \second{\meanstd{48.6}{2.6}} & \second{\meanstd{44.1}{3.8}} \\
        \bottomrule
    \end{tabular}
\end{table*}

EMPA evaluates persona-aligned empathy at the trajectory level through EPM-Idx, which aggregates Outcome (RDI, Etot, Snet), Efficiency (Rho, Sproj, Tau), and Stability (Rpos, Align, Pen), with all reported indices normalized so that higher is better. As shown in Table~\ref{tab:empa-outcome-full}, \oursrlver{} achieves the highest EPM-Idx under all three judges, scoring $95.77$, $83.54$, and $38.44$ under DeepSeek-V4-Pro, Gemini-2.5-Pro, and GPT-5.5, respectively, corresponding to gains of $43.67$, $55.43$, and $13.02$ over reproduced MICA. The largest improvements occur in Outcome, particularly RDI, Etot, and Snet, indicating stronger final psychological progress and greater cumulative effective support, while \ourmethod{} also consistently improves Efficiency and Stability over MICA.

\section{Human Preference Evaluation}
\label{app:human_eval}

To complement the automatic evaluations, we conduct a blinded human preference study comparing the end-to-end response quality of \oursrlver, MICA, and RLVER.
We randomly sample 50 dialogue contexts from the 166 held-out examples used for rubric-generator evaluation in \S\ref{sec:ablation}.
For each example, we discard the original chosen and rejected responses and retain only the dialogue context ending with the final user turn.
Given the same context, the three models independently generate responses to the final user message.

We recruit 15 annotators with backgrounds in psychology.
Each annotator is randomly assigned 30 contexts from the 50-context pool.
For each context, the annotator is shown the dialogue history together with three anonymized candidate responses and asked to select the response they prefer overall.
Annotators may select a tie when no response is clearly preferred.
Model identities are hidden throughout the evaluation, and response order is randomly shuffled to mitigate position bias.

We report the \emph{human preference rate} after excluding tie judgments, computed as the fraction of remaining selections received by each model.
As shown in Table~\ref{tab:human_preference}, \oursrlver{} receives 51.5\% of human selections, compared with 23.9\% for MICA and 24.6\% for RLVER.
Thus, among judgments with a clear preference, more than half favor responses generated by \oursrlver.
These results provide complementary human evidence that the gains from context-adaptive rubrics are also reflected in direct preferences over generated responses.

\begin{table}[t]
    \centering
    \caption{
        Blinded human preference evaluation.
        Fifteen annotators each evaluate 30 randomly assigned contexts.
        Model identities are hidden and response order is randomly shuffled.
    }
    \label{tab:human_preference}
    \small
    \setlength{\tabcolsep}{5pt}
    \begin{tabular}{@{}>{\centering\arraybackslash}p{0.43\linewidth}
                        >{\centering\arraybackslash}p{0.22\linewidth}
                        >{\centering\arraybackslash}p{0.28\linewidth}@{}}
        \toprule
        \textbf{Model} &
        \textbf{ Selections} &
        \textbf{Preference (\%) $\uparrow$} \\
        \midrule
        \maporep & 104 & 23.9 \\
        \rlver & 107 & 24.6 \\
        \rowcolor{ourscyan}\oursrlver & 224 & \textbf{51.5} \\
        \bottomrule
    \end{tabular}
\end{table}
\section{Strategy-to-dimension mapping}
\label{app:strategy-dimension-mapping}

We harmonize the strategy annotations in ESConv and MINT by mapping them into the three empathy dimensions used in our rubric framework: cognitive, affective, and proactive empathy. The mapping is based on the primary function of each strategy. Strategies that facilitate understanding or interpretation of the seeker’s situation are categorized as cognitive empathy; those that convey emotional attunement, validation, or interpersonal support are categorized as affective empathy; and those that provide actionable guidance or practical support are categorized as proactive empathy. Table~\ref{tab:strategy-dimension-mapping} summarizes the resulting mappings for both datasets.

\begin{table}[H]
    \centering
    \caption{Mapping from ESConv strategies and MINT strategies to empathy dimensions.}
    \label{tab:strategy-dimension-mapping}
    \small
    \setlength{\tabcolsep}{5pt}
    \begin{tabular}{@{}>{\centering\arraybackslash}p{0.16\linewidth}>{\raggedright\arraybackslash}p{0.37\linewidth}>{\raggedright\arraybackslash}p{0.39\linewidth}@{}}
        \toprule
        \textbf{Dimension} & \textbf{ESConv strategies} & \textbf{MINT strategies} \\
        \midrule
        Cognitive &
        Question; Restatement or Paraphrasing; Others &
        questioning; paraphrasing; contextualizing; reappraisal \\
        \addlinespace[0.35ex]
        Affective &
        Reflection of feelings; Affirmation and Reassurance; Self-disclosure &
        validation; emotional\_expression; terms\_of\_endearment; self\_disclosure; solidarity; gratitude; spirituality \\
        \addlinespace[0.35ex]
        Proactive &
        Providing Suggestions; Information &
        advice; information; empowerment; assistance \\
        \bottomrule
    \end{tabular}
\end{table}

\section{SFT data construction}
\label{app:sft-data}

For rubric-generator SFT, we first filter the source dialogues from MINT and ESConv by topic, retaining only emotion-related conversational tasks. We then map the original empathy strategies to the three CARE empathy dimensions according to the mapping rules in \S~\ref{app:strategy-dimension-mapping}. The dialogue history, current query, reference response, original strategies, and mapped dimensions are provided to DeepSeek-v4-pro to generate turn-specific rubrics, using the prompt provided in \S~\ref{app:prompts}. After format validation, each candidate rubric is independently evaluated by two judges: Qwen3.6-27B and gemma-4-31B.Each judge evaluates the three empathy categories in terms of turn specificity, criterion clarity, and within-category non-redundancy. Separately, the judge scores the original supporter response from 0 to 5 on each criterion, and these scores are aggregated using the category weights to obtain an overall score. A rubric is accepted only if both judges pass all three category-level quality checks and their overall scores differ by no more than 0.2 points. The final SFT corpus contains 1.5k cleaned examples, each with an average of 8.63 dialogue turns. The resulting rubrics contain an average of 7.67 criteria and 970.72 words, with each criterion description averaging 102.72 words. The two examples below show parsed SFT records in JSON format.

\begin{promptbox}{SFT Data Example 1}
{
  "messages": [
    {
      "role": "system",
      "content": "You are a rubric generator. Given the dialogue history up to the seeker's latest turn, output a JSON rubric evaluating the upcoming supporter response. Output JSON only, no explanation.\nSchema: {\"categories\":[{\"category\",\"weight_rationale\",\"category_weight\",\"criteria\":[{\"title\",\"description\"}]}]}\n- \"category\" must be exactly one of: Cognitive Empathy, Affective Empathy, Proactive Empathy.\n- \"weight_rationale\" is one concise sentence, written BEFORE category_weight, tying the seeker's latest-turn need and the supporter's empathy tactics to this category's weight (state whether it is the highest, lowest, or tied).\n- Escape any double quotes appearing inside string values."
    },
    {
      "role": "user",
      "content": "
      seeker: What should I do if I want to know a person's real thoughts but we don't always have chance to meet and he's unwilling to be taken to a more privacy place

    supporter: As an AI language model, it is not appropriate for me to provide advice on how to invade someone's privacy or extract their thoughts without their consent. It's important to respect people's boundaries and decisions. If you want to get to know someone better, try to establish a rapport and build a trusting relationship with them over time. You can also ask open-ended questions and listen actively to their responses to get a better understanding of their thoughts and feelings. It's essential to have mutual respect and trust in any relationship.
    
    seeker: You may get some misunderstanding. We are friends and we haven't seen each other for a long time. He asked me to have a short connection with him and when I do so, he just keeps silent. I think he may have some words to say but I don't know how to...
    
    supporter: I see. In this case, it may be helpful to communicate your feelings and intentions directly with your friend. You can try to reach out to him and express that you are interested in catching up and hearing his thoughts and feelings. It's important to approach the situation with empathy and sensitivity, and to respect his boundaries if he is not comfortable sharing certain things with you. You can also try to schedule a time to meet up in person or over video chat to have a more meaningful and focused conversation. Remember to listen actively and be present in the moment, and not to pressure him to share anything he is not comfortable with.
    
    seeker: But I think it's a little strange if I just reach out for him and be direct suddenly
      "
    },
    {
      "role": "assistant",
      "content": {
        "categories": [
          {
            "category": "Cognitive Empathy",
            "weight_rationale": "This is the highest-weighted category because the supporter's primary task is to accurately interpret the seeker's explicit dilemma (feeling strange about being direct) and provide a coherent, contextually-grounded strategy for re-engagement.",
            "category_weight": 0.4,
            "criteria": [
              {
                "title": "Accurately Understand the Core Dilemma",
                "description": "[0] Completely misinterprets the seeker's concern. [1] Addresses a different problem entirely. [2] Partially grasps the concern but misses the central conflict between wanting to connect and feeling awkward. [3] Correctly identifies the core dilemma of feeling hesitant to be direct. [4] Accurately captures the dilemma and the underlying desire for a natural reconnection. [5] Precisely articulates the nuanced tension between the seeker's goal and their social discomfort. value: 0, 1, 2, 3, 4, 5"
              },
              {
                "title": "Provide Coherent and Relevant Reasoning",
                "description": "[0] Response is nonsensical or completely off-topic. [1] Reasoning is illogical and disconnected from the seeker's situation. [2] Offers some relevant points but the logic is disjointed or contains a significant tangent. [3] Presents a clear, logical sequence of points that directly addresses the seeker's hesitation. [4] Builds a compelling and well-structured argument that connects the seeker's feeling to a practical path forward. [5] Reasoning is exceptionally coherent, weaving the seeker's emotional state and the proposed action into a seamless, insightful narrative. value: 0, 1, 2, 3, 4, 5"
              },
              {
                "title": "Stay Grounded in the Seeker's Context",
                "description": "[0] Ignores the described friendship and history. [1] Gives generic advice irrelevant to reconnecting with an old friend. [2] Mentions the friendship but advice is mostly generic. [3] Advice is clearly tailored to the context of a long-lost friend and a desire for a non-awkward reconnection. [4] Explicitly references the seeker's specific situation (long time no see, friend's silence) to justify the suggested approach. [5] Masterfully integrates every detail of the seeker's context to make the advice feel uniquely personal and perfectly fitted. value: 0, 1, 2, 3, 4, 5"
              }
            ]
          },
          {
            "category": "Affective Empathy",
            "weight_rationale": "This category is weighted lowest because the turn's primary focus is on providing a cognitive strategy and actionable steps. While emotional validation is a necessary foundation, the bulk of the response is dedicated to problem-solving.",
            "category_weight": 0.25,
            "criteria": [
              {
                "title": "Validate the Feeling of Awkwardness",
                "description": "[0] Dismisses or ridicules the seeker's feeling. [1] Ignores the expressed hesitation entirely. [2] Acknowledges the feeling but in a perfunctory or dismissive way. [3] Explicitly validates that feeling hesitant or awkward is understandable in this situation. [4] Normalizes the feeling by connecting it to the universal experience of reconnecting after a long time. [5] Deeply validates the emotion, making the seeker feel truly seen and accepted for their vulnerability before offering any advice. value: 0, 1, 2, 3, 4, 5"
              },
              {
                "title": "Convey a Warm and Encouraging Tone",
                "description": "[0] Tone is cold, judgmental, or robotic. [1] Tone is neutral and detached, offering no emotional support. [2] Attempts a supportive tone but comes across as slightly patronizing or formulaic. [3] Maintains a consistently warm, respectful, and non-judgmental tone. [4] Tone is genuinely encouraging and empathetic, fostering a sense of safety and hope. [5] The language is so warmly human and supportive that it feels like a trusted friend providing comfort and courage. value: 0, 1, 2, 3, 4, 5"
              }
            ]
          },
          {
            "category": "Proactive Empathy",
            "weight_rationale": "This category is weighted second-highest because the core of the supporter's turn is to move the conversation forward by offering a concrete, actionable plan. The response's usefulness hinges on the quality and appropriateness of this proactive guidance.",
            "category_weight": 0.35,
            "criteria": [
              {
                "title": "Offer an Appropriate and Low-Pressure Strategy",
                "description": "[0] Suggests a harmful, aggressive, or completely unfeasible action. [1] Provides a vague or unhelpful suggestion that doesn't solve the awkwardness problem. [2] Suggests a reasonable action but fails to address the seeker's need for a non-awkward approach. [3] Provides a concrete, feasible strategy (like sending a casual message) that directly mitigates the fear of being too direct. [4] The strategy is not only feasible but cleverly designed to feel natural and low-pressure, perfectly aligning with the seeker's comfort level. [5] Offers a brilliantly nuanced, multi-step approach that empowers the seeker to initiate contact with complete confidence and zero pressure. value: 0, 1, 2, 3, 4, 5"
              },
              {
                "title": "Provide Actionable Conversational Tools",
                "description": "[0] Provides no examples or tools. [1] Gives abstract advice without any practical application. [2] Mentions a type of tool (e.g., 'ask open-ended questions') but doesn't make it usable. [3] Provides specific, concrete examples of what to say or ask that the seeker can immediately use. [4] The examples are highly tailored to the context of reconnecting with an old friend and are easy to adapt. [5] Provides a versatile toolkit of conversational openers and explains the rationale behind them, empowering the seeker to generate their own. value: 0, 1, 2, 3, 4, 5"
              },
              {
                "title": "Respect the Seeker's Autonomy and Pace",
                "description": "[0] Commands the seeker to take a specific action. [1] Pressures the seeker to act despite their stated discomfort. [2] Frames the advice as a suggestion but with an underlying tone of 'you must do this.' [3] Clearly frames the proposed actions as optional suggestions, respecting the seeker's choice. [4] Explicitly empowers the seeker to decide what feels right for them, reinforcing their control over the situation. [5] Beautifully balances encouragement with complete respect for the seeker's autonomy, making it clear that any action is entirely their choice and on their timeline. value: 0, 1, 2, 3, 4, 5"
              }
            ]
          }
        ]
      }
    }
  ]
}
\end{promptbox}

\begin{promptbox}{SFT Data Example 2}
{
  "messages": [
    {
      "role": "system",
      "content": "You are a rubric generator. Given the dialogue history up to the seeker's latest turn, output a JSON rubric evaluating the upcoming supporter response. Output JSON only, no explanation.\nSchema: {\"categories\":[{\"category\",\"weight_rationale\",\"category_weight\",\"criteria\":[{\"title\",\"description\"}]}]}\n- \"category\" must be exactly one of: Cognitive Empathy, Affective Empathy, Proactive Empathy.\n- \"weight_rationale\" is one concise sentence, written BEFORE category_weight, tying the seeker's latest-turn need and the supporter's empathy tactics to this category's weight (state whether it is the highest, lowest, or tied).\n- Escape any double quotes appearing inside string values."
    },
    {
      "content": "seeker: I am having a conflict with my peers around the process and
    prioritzation of operational symmetry across a [TEAM] of [NUMBER] people.
    
    I need to understand why this is so important to them at this time, and help
    express the additional tax this is going to take on my folks. It feels like
    \"they have more time\" to think about this, whereas I feel like my team is more
    focused on \"product quality.\"
    
    supporter: I understand that you're in a challenging situation dealing with
    conflicts around process and prioritization. It sounds like your peers may be
    emphasizing operational symmetry, while you're focused on maintaining product
    quality, which creates tension.
    
    This discrepancy appears to be placing stress on you and your team. It can be
    so hard when it feels like your team is overstretched, and it seems as though
    others may not fully understand your perspective.
    
    May I suggest that you consider initiating a dialogue with your peers? It could
    create an opportunity for you to express your concerns, and for them to explain
    why operational symmetry is so important to them right now.
    
    This may facilitate a better understanding of their perspective and could allow
    for a collaborative conversation around how to prioritize the needs of both
    parties.
    
    Talking about the strain on your team could also make it clear to them the
    additional burden their requirements could impose. Phrase it from a place of
    wanting to find a solution beneficial to all.
    
    Something along the lines of, \"We value operational symmetry and see its
    importance. However, our current focus heavily leans on ensuring product
    quality. Can we find a strategy that respects both needs?\"
    
    Remember, you are all working towards a common goal - the success of the studio,
    and having different viewpoints can often lead to creative and more robust
    solutions when managed well.
    
    It's important to use these conversations not just as a way to voice your
    concerns but also to mutually understand and respect each other's priorities.
    
    Let me know how it goes, or if there's anything else that you need help with in
    managing this situation.
    
    seeker: Thanks, those are good ideas. I'm a little behind and catching up, and
    feel like the rest of my peers are already on board for this approach, including
    my manager.
    
    I have expressed my understanding of the importance of operations, and that my
    team is interested and has a lot on their plates. I'm starting to feel a bit
    othered.
    
    supporter: I'm really sorry to hear that you're feeling this way, it must be
    quite challenging for you.
    
    It seems as though you're working incredibly hard to maintain the balance
    between operational priorities and the workload of your own team, all the while
    feeling somewhat isolated among your peers and your manager.
    
    That's a tough position to be in.
    
    First, allow yourself a moment to recognize your efforts. It's clear you're
    shouldering a lot of pressure and it's crucial not to lose sight of your value
    and all the hard work you're putting into trying to find a solution for everyone
    involved.
    
    As for feeling 'othered', it's really important to address this before it
    starts to affect your motivation and mental well-being.
    
    You might try requesting a private discussion with your manager, where you can
    candidly express your feelings about your current situation.
    
    Perhaps something like, \"I realize that everyone is on board with pushing
    operational symmetry, and I respect that. I just feel a bit isolated from the
    group because of my current commitments and obligations to our team.
    
    Could we discuss potential ways to bridge this gap so that myself and my team
    can effectively contribute without feeling overwhelmed?\"
    
    They might be able to provide reassurance, offer suggestions, or even help to
    facilitate conversations with your peers to work out a compromise or an updated
    plan that takes your team's workload into account.
    
    Dealing with complexities and pressures like this isn't easy. But through
    effective communication, you are taking a proactive approach towards finding a
    resolution.
    
    Remember, it's okay to ask for help when you need it. You're not alone in this.
    
    Let me know if there's anything more I can do, or if you want to explore other
    strategies for addressing this feeling of disconnection.
    
    seeker: It's especially hard because I feel like they've already decided
    \"this is the way we're going to do this.\"
    
    supporter: I'm sorry you're feeling that way. It's certainly challenging to feel
    like decisions have been made without considering the full impact on your team's
    workload.
    
    It must feel as though your team's focus on product quality is not being
    adequately considered, and that can create a sense of being overlooked or
    marginalized.
    
    While it might seem that the decision is already made and there is a lack of
    room for flexibility, it could be beneficial to voice your concerns and
    perspectives in a constructive way if you haven't already.
    
    More often than not, people don't realize the full impact of their decisions on
    others until it is openly communicated to them in a considerate manner.
    
    Consider constructing your thoughts in a clear, concise proposal for your
    manager and peers. This proposal could detail your challenges and offer
    suggestions for 'how' it can be implemented to consider everyone's needs.
    
    That could potentially open up the floor for discussion again.
    
    Also, remember that it is okay to ask for ongoing support throughout the
    implementation. Perhaps there could be a phased approach, or additional
    resources provided, to ease the transition and workload for your team?
    
    What counts the most is that you are voicing your concerns for your team and
    standing up for what you think is important for them and the quality of your
    work. That is admirable.
    
    Please know that these situations can be tough to navigate, but you're taking
    significant strides by seeking advice and showing readiness to address the
    situation.
    
    Let me know if you'd like help brainstorming more ideas or if you need support
    in other ways. We're here to help you through this.
    
    seeker: I have one person on my team who already has a process that is working
    for them.
    
    They don't want to have to make changes \"just because\" the larger team wants
    to operate similiarly."
    },
    {
      "role": "assistant",
      "content": {
        "categories": [
          {
            "category": "Cognitive Empathy",
            "weight_rationale": "This turn requires high cognitive empathy to accurately frame the multi-layered problem (peer conflict + team resistance) and provide a coherent, logical strategy for the conversation with the resistant team member.",
            "category_weight": 0.4,
            "criteria": [
              {
                "title": "Accurately Frame the Multifaceted Tension",
                "description": "[0] Completely misrepresents the core conflict, ignoring either the peer pressure or the team member's resistance. [1] Acknowledges only one side of the problem, missing the tension between external demands and internal resistance. [2] Mentions both sides but fails to connect them as a single, multifaceted challenge for the seeker. [3] Clearly identifies the tension between the team member's resistance and the need for operational symmetry, framing it as a difficult balance. [4] Insightfully frames the tension, explicitly linking the team member's resistance to the seeker's earlier feelings of being 'othered' by peers. [5] Masterfully synthesizes the entire context, framing the issue as a complex system where the team member's resistance is a microcosm of the seeker's own struggle for autonomy and validation. value: 0, 1, 2, 3, 4, 5"
              },
              {
                "title": "Understand the Resistant Team Member's Perspective",
                "description": "[0] Dismisses or ignores the team member's perspective, focusing only on enforcing the new process. [1] Acknowledges the team member has a working process but labels the resistance as simple stubbornness. [2] Recognizes the resistance stems from a perception of unnecessary change but doesn't explore the underlying reasons. [3] Demonstrates understanding that the resistance is rooted in a valid, efficient existing process and a natural aversion to change 'just because.' [4] Deeply explores the specific, unspoken concerns behind the resistance, such as fear of reduced efficiency or loss of autonomy. [5] Anticipates and articulates the team member's core values and identity tied to their process, framing the change as a potential threat to their sense of competence. value: 0, 1, 2, 3, 4, 5"
              },
              {
                "title": "Provide a Coherent Conversational Strategy",
                "description": "[0] Offers no strategy or a completely illogical sequence of steps. [1] Suggests a vague or counterproductive action, like immediately demanding compliance. [2] Proposes a basic sequence (e.g., 'talk to them') but lacks a clear, reasoned flow. [3] Outlines a logical, step-by-step approach: first explore concerns, then explain benefits tailored to those concerns. [4] Provides a nuanced strategy that sequences empathy, inquiry, and tailored persuasion, anticipating how each step builds on the last. [5] Delivers a masterful, psychologically astute strategy that not only sequences actions but also predicts the team member's likely reactions and prepares the seeker for a collaborative dialogue. value: 0, 1, 2, 3, 4, 5"
              }
            ]
          },
          {
            "category": "Affective Empathy",
            "weight_rationale": "Affective empathy is important to validate the seeker's compounded stress and the team member's discomfort, but the primary need is for actionable cognitive and proactive support to resolve the situation.",
            "category_weight": 0.25,
            "criteria": [
              {
                "title": "Validate the Seeker's Compounded Challenge",
                "description": "[0] Invalidates the seeker's feelings, blaming them for the situation. [1] Offers a generic platitude without connecting to the specific difficulty of managing a resistant team member while feeling isolated. [2] Acknowledges the situation is 'challenging' but doesn't name the specific emotional toll of balancing everyone's needs. [3] Explicitly validates the difficulty of balancing the team member's needs, peer expectations, and personal feelings of being 'othered,' naming it as a heavy burden. [4] Resonates with the seeker's feeling of being caught in the middle, validating the loneliness and pressure of advocating for their team while feeling unsupported themselves. [5] Profoundly mirrors the seeker's emotional exhaustion and sense of isolation, making them feel deeply seen not just as a manager, but as a person navigating a painful, no-win situation. value: 0, 1, 2, 3, 4, 5"
              },
              {
                "title": "Acknowledge the Team Member's Emotional Reality",
                "description": "[0] Shows no recognition of the team member's potential feelings, treating them as an obstacle. [1] Acknowledges the team member might be 'uncomfortable' but frames it as an overreaction. [2] Recognizes the discomfort of change but doesn't connect it to the specific context of a process that already works. [3] Validates that change is genuinely challenging, especially when a current process is efficient, making the resistance understandable. [4] Empathizes with the team member's potential frustration and feeling that their expertise is being undervalued by a top-down decision. [5] Conveys deep emotional attunement to the team member's likely sense of pride and ownership over their work, validating the change as a personal and professional disruption. value: 0, 1, 2, 3, 4, 5"
              }
            ]
          },
          {
            "category": "Proactive Empathy",
            "weight_rationale": "Proactive empathy is equally critical as cognitive empathy here, as the seeker needs concrete, actionable steps to navigate the conversation with their team member and move from feeling stuck to feeling empowered.",
            "category_weight": 0.35,
            "criteria": [
              {
                "title": "Offer Actionable Conversation Tactics",
                "description": "[0] Provides no actionable advice or suggests harmful actions. [1] Gives vague advice like 'be supportive' without any concrete conversational tools. [2] Suggests a general approach (e.g., 'explain the benefits') but lacks specific, usable phrasing or techniques. [3] Provides concrete, actionable tactics, such as framing the conversation around the team member's specific concerns and using 'what' and 'how' questions to explore resistance. [4] Offers detailed conversational frameworks, including example openers and methods to collaboratively problem-solve, directly addressing the 'just because' sentiment. [5] Delivers a comprehensive, turnkey conversational plan with multiple tailored options, fallback strategies, and guidance on reading the team member's cues to adapt in real-time. value: 0, 1, 2, 3, 4, 5"
              },
              {
                "title": "Empower the Seeker's Leadership Role",
                "description": "[0] Undermines the seeker's authority or suggests they are failing as a leader. [1] Offers support but fails to reinforce the seeker's capability or agency. [2] Makes a generic statement about the seeker doing a good job without tying it to their specific actions. [3] Explicitly affirms the seeker's commendable action of considering individual team needs, reinforcing their thoughtful leadership. [4] Actively empowers the seeker by framing their advocacy for the team member as a crucial leadership strength, not an obstacle. [5] Transforms the seeker's perspective, helping them see this challenge as an opportunity to model inclusive leadership and build a stronger, more trusting team culture. value: 0, 1, 2, 3, 4, 5"
              },
              {
                "title": "Guide the Pace of Change and Support",
                "description": "[0] Pushes for immediate, forceful implementation of the new process. [1] Acknowledges change takes time but offers no practical guidance on pacing. [2] Mentions patience but doesn't connect it to a concrete, phased approach. [3] Advises patience and continued support, explicitly linking the need for adjustment time to the process of helping the team member see the 'why.' [4] Recommends a phased or incremental approach to introducing the change, with checkpoints to reassess the team member's comfort and gather feedback. [5] Outlines a sophisticated change-management micro-strategy, including celebrating small wins, creating psychological safety for feedback, and normalizing the adjustment period as part of the process. value: 0, 1, 2, 3, 4, 5"
              }
            ]
          }
        ]
      }
    }
  ]
}
\end{promptbox}

\section{Preference data construction}
\label{app:preference-data}

For rubric-generator RL, we first collect multi-turn emotional-support dialogues between human users and dialogue models, use intermediate turns as fixed dialogue states, and roll out GPT-5, Gemini-3-Pro-Preview, and DeepSeek-V3.2 from the same preceding context to obtain comparable candidate responses. A disjoint annotator group then evaluates one anonymized response pair at a time in randomized order; each pair is judged by three annotators, and the majority decision gives the preference label for preference RL. For the majority-preferred responses, we further obtain coarse support-strategy labels with the MINT annotator and map them to empathy dimensions using the fixed mapping in \S~\ref{app:strategy-dimension-mapping}. The preference data used to train the rubric generator contains 2.8k pairs.  On average, each example contains 12.42 turns; seeker turns average 31.04 words, while the chosen and rejected supporter responses average 75.78 and 70.00 words, respectively. The two examples below show parsed preference records in JSON format.

\begin{promptbox}{Preference Data Example 1}
{
  "prompt": [
    {
      "role": "system",
      "content": "You are a rubric generator. Given the dialogue history up to the seeker's latest turn, output a JSON rubric evaluating the upcoming supporter response. Output JSON only, no explanation. Schema: {\"categories\":[{\"category\",\"weight_rationale\",\"category_weight\",\"criteria\":[{\"title\" ,\"description\"}]}]} - \"category\" must be exactly one of: Cognitive Empathy, Affective Empathy, Proactive Empathy. - \"weight_rationale\" is one concise sentence, written BEFORE category_weight, tying the seeker's latest-turn need and the supporter's empathy tactics to this category's weight (state whether it is the highest, lowest, or tied). - Escape any double quotes appearing inside string values."
    },
    {
      "role": "user",
      "content": [
        "seeker: Sometimes I think, I really should learn to love myself more.",
        "supporter: Loving yourself really is important. The moment you start having that thought, you're already moving closer to a better version of yourself. Are there any moments in your daily life where you feel like you haven't been treating yourself well?",
        "seeker: I just want to buy some things I've been wanting. I know some people might say it's all just a capitalist trap, but I don't think there's any harm in buying one or two things I like.",
        "supporter: I think you're right. Treating yourself to a couple of things you like now and then is perfectly fine, as long as it's within your budget and makes you happy. It's really a form of self-care spending, an investment in yourself that can boost your happiness. Is there anything in particular you've been wanting to buy lately?",
        "seeker: There are quite a few things I've been wanting to buy lately. For example, I want to get a game cartridge.",
        "supporter: Buying a game you love can definitely help you unwind. Since you mentioned getting a cartridge, do you usually play on a Switch, PS5, or some other handheld?",
        "seeker: I didn't mean going to a physical store to buy it. I think just buying it online is fine; going out to a store is such a hassle."
      ]
    }
  ],
  "extra_info": {
    "chosen_text": "Exactly, shopping online means you don't have to go out, saving time and energy, which is perfect for when you just want to relax. So, are you looking for a specific game, or do you want to see what everyone else is recommending?",
    "rejected_text": "Buying online is definitely convenient. You can find them on platforms like Taobao or Xianyu, both brand new and second-hand. Are you looking for a newly released game, or a classic older one?",
    "chosen_tactics": [
      "questioning",
      "information"
    ]
  }
}
\end{promptbox}

\begin{promptbox}{Preference Data Example 2}
{
  "prompt": [
    {
      "role": "system",
      "content": "You are a rubric generator. Given the dialogue history up to the seeker's latest turn, output a JSON rubric evaluating the upcoming supporter response. Output JSON only, no explanation. Schema: {\"categories\":[{\"category\",\"weight_rationale\",\"category_weight\",\"criteria\":[{\"title\" ,\"description\"}]}]} - \"category\" must be exactly one of: Cognitive Empathy, Affective Empathy, Proactive Empathy. - \"weight_rationale\" is one concise sentence, written BEFORE category_weight, tying the seeker's latest-turn need and the supporter's empathy tactics to this category's weight (state whether it is the highest, lowest, or tied). - Escape any double quotes appearing inside string values."
    },
    {
      "role": "user",
      "content": [
        "seeker: My friend, he absolutely hates his mom, but he can't leave her. After all, he can't really grasp what home and family are supposed to mean, and there's no place left for him in this world.",
        "supporter: That feeling of having nowhere to go is truly heartbreaking to hear. Hating her, yet being forced to depend on her--this kind of conflicted relationship must feel like a cage trapping him. You said he can't understand home and family; is it because he's never truly received acceptance or a sense of safety at home?",
        "seeker: Exactly. Right now, he doesn't know what love is, or what it means to be loved. He's completely clueless about all of it, and he doesn't even know if he should have been born.",
        "supporter: When someone starts questioning the very meaning of their own birth, it shows just how deep the emotional wounds go. In a place that should be a safe harbor, he never learned what love is--that kind of absence can leave a person feeling profoundly empty and lost. As his friend, seeing him deny his own worth like this must be really painful for you too.",
        "seeker: It might be because, for a long time, his mom has been, um, neglectful or critical, or even lashing out at him with insults for no reason.",
        "supporter: The neglect makes him feel like he doesn't matter, and the criticism and insults constantly tear down his self-esteem. Living under that kind of verbal abuse for so long, it's easy for a person to develop learned helplessness, feeling like everything they do is wrong. No wonder he feels unworthy of love--because even the person closest to him keeps denying him.",
        "seeker: I think there's definitely something going on inside him psychologically, so he really needs a therapist to help him work through it."
      ]
    }
  ],
  "extra_info": {
    "chosen_text": "Your judgment is right--this kind of long-term psychological trauma really does need professional intervention to help him rebuild his understanding of himself. A therapist can provide a safe environment where he can release all those pent-up emotions. But considering how entangled his relationship with his mom is, do you think he's in a position right now, or even has the willingness, to actually go to therapy?",
    "rejected_text": "Yes, being in that kind of environment for so long, it's inevitable that a lot of trauma builds up inside. Professional guidance can help him see that the problem isn't his fault. But is he willing to go himself? Or have you ever tried bringing up this suggestion with him?",
    "chosen_tactics": [
      "paraphrasing",
      "questioning",
      "information",
      "validation",
      "advice"
    ]
  }
}
\end{promptbox}

\section{Prompts}
\label{app:prompts}

\begin{promptbox}{SFT Rubric Tagger Prompt}
Generate one turn-specific empathy-evaluation rubric. Return ONLY JSON with top-level keys `analysis` and `rubric`.

Use all inputs: conversation context, internal calibration response, tactic signals, schema example, and validator feedback if present.

Output contract:
- `analysis`: non-empty `user_state`, `expected_supporter_action`, `local_context_need`, `quality_risks`, `calibration_sanity_check`.
- `rubric.categories`: exactly Cognitive Empathy, Affective Empathy, Proactive Empathy, in that order.
- Each category: `weight_rationale`, then `category_weight`, then 2-4 criteria.
- Each criterion: only `title` + `description`; description includes `[0]`..`[5]` and ends with `value: 0, 1, 2, 3, 4, 5`.

Quality guidance:
1. Weights sum to 1.0 and should broadly follow `category_weight_tiers`.
2. Criteria counts should reflect this turn's tactic mix and actual support needs.
3. `weight_rationale` should be consistent with its weight without overclaiming rank.
4. Avoid prompt/meta words in rubric text: gold, reference, ideal, ground truth, target response/answer/reply.
5. Avoid answer leakage: do not copy, quote, paraphrase, summarize, or reveal distinctive wording, examples, advice items, named resources, structure, or closing sentences from the calibration response. Score levels must describe abstract qualities, not answer content.

Retry repair:
If `previous_failed_attempt_feedback_MUST_FIX` exists, fix it before anything else. Repair exact schema, criteria count, level-format, duplicate-title, category, or weight-sum errors named by the validator.

Regeneration controls:
If `locked_categories_to_preserve` is non-empty, keep those category objects exactly and regenerate only `categories_to_regenerate`.

Final self-check: JSON shape valid; no copied calibration phrase; no forbidden prompt words; rationale ranks match weights; criteria counts match tactic coverage.
\end{promptbox}

\begin{promptbox}{SFT Data Validation Judge Prompt}
You are a strict judge for an empathy rubric and the gold_supporter_response. Return only valid JSON with exactly these keys: category_quality_checks, criterion_scores, revision_advice.

category_quality_checks must be a JSON object whose keys are the exact category names (NOT a list): Affective Empathy, Cognitive Empathy, Proactive Empathy. Each category maps to an object with exactly these boolean keys: criteria_clear, non_redundant, turn_specific (turn_specific: the category's rubric fits this exact turn; criteria_clear: criteria are clear and usable; non_redundant: criteria within the category are not duplicative).

criterion_scores must score the gold_supporter_response against the candidate rubric: one item per rubric criterion with category, criterion_title, and score (NO rationale). Use each criterion's own category and title exactly as given in candidate_rubric. Each score is an integer in {0, 1, 2, 3, 4, 5} where higher means stronger performance on that criterion (0 = absent/poor, 5 = excellent). Do not output a total. revision_advice must be one short string. Follow required_output_shape exactly.
\end{promptbox}

\begin{promptbox}{Inference-Time Rubric-Generator Prompt}
You are a rubric generator. Given the dialogue history up to the seeker's latest turn, output a JSON rubric evaluating the upcoming supporter response. Output JSON only, no explanation.

Schema: {"categories":[{"category","weight_rationale","category_weight","criteria":[{"title","description"}]}]}

- "category" must be exactly one of: Cognitive Empathy, Affective Empathy, Proactive Empathy.
- "weight_rationale" is one concise sentence, written BEFORE category_weight, tying the seeker's latest-turn need and the supporter's empathy tactics to this category's weight (state whether it is the highest, lowest, or tied).
- Escape any double quotes appearing inside string values.
\end{promptbox}

\end{document}